\documentclass[10pt,journal,compsoc]{IEEEtran}

\ifCLASSOPTIONcompsoc
  \usepackage[nocompress]{cite}
\else
  \usepackage{cite}
\fi

\usepackage{cite}
\usepackage{graphicx}
\usepackage{url}
\usepackage{amsmath}
\usepackage{amssymb}
\usepackage[caption = false, font=footnotesize]{subfig}
\usepackage{float}
\usepackage{booktabs} 
\usepackage{multirow}
\usepackage[colorlinks=true,linkcolor=blue]{hyperref}%
\usepackage{array,ulem}
\usepackage{color}
\usepackage{tabularx}
\usepackage{longtable}
\usepackage{tabu}
\usepackage{pifont}
\usepackage{amssymb}
\usepackage{ragged2e}

\begin{document}

\title{Generative Models at the Frontier of Compression: A Survey on Generative Face Video Coding}

\author{Bolin Chen,~\IEEEmembership{Member,~IEEE}, 
        Shanzhi Yin, 
        Goluck Konuko,
        Giuseppe Valenzise,~\IEEEmembership{Senior Member,~IEEE},         
        Zihan Zhang,
        Shiqi Wang,~\IEEEmembership{Senior Member,~IEEE},        
        Yan Ye,~\IEEEmembership{Senior Member,~IEEE}
\IEEEcompsocitemizethanks{
\IEEEcompsocthanksitem The first three authors (Bolin Chen, Shanzhi Yin and Goluck Konuko) contributed equally to this work and should be regarded as co-first authors.\\
\IEEEcompsocthanksitem Bolin Chen is with DAMO Academy, Alibaba Group, Hangzhou, 310023, China and Hupan Lab, Hangzhou, 310023, China (e-mail: chenbolin.chenboli@alibaba-inc.com).\\
\IEEEcompsocthanksitem Shanzhi Yin, Zihan Zhang and Shiqi Wang are with the Department of Computer Science, City University of Hong Kong, Hong Kong (e-mail: shanzhyin3-c@my.cityu.edu.hk, zhzhang38-c@my.cityu.edu.hk and shiqwang@cityu.edu.hk). \\
\IEEEcompsocthanksitem Goluck Konuko was at L2S - CentraleSup\`elec and LTCI - Telécom Paris for the duration of this work (e-mail: goluck.konuko@centralesupelec.fr)\\
\IEEEcompsocthanksitem Giuseppe Valenzise is with the CNRS, CentraleSupelec, Laboratoire desSignaux et Systèmes, Université Paris-Saclay, 91190 Gif-sur-Yvette, France (e-mail: giuseppe.valenzise@l2s.centralesupelec.fr).\\
\IEEEcompsocthanksitem Yan Ye is with the Alibaba Group, Sunnyvale, CA, USA and Hupan Lab, Hangzhou, 310023, China (e-mail: yan.ye@alibaba-inc.com).}
}

\IEEEtitleabstractindextext{%
\begin{abstract}
\justifying
The rise of deep generative models has greatly advanced video compression, reshaping the paradigm of face video coding through their powerful capability for semantic-aware representation and lifelike synthesis. Generative Face Video Coding (GFVC) stands at the forefront of this revolution, which could characterize complex facial dynamics into compact latent codes for bitstream compactness at the encoder side and leverages powerful deep generative models to reconstruct high-fidelity face signal from the compressed latent codes at the decoder side. As such, this well-designed GFVC paradigm could enable high-fidelity face video communication at ultra-low bitrate ranges, far surpassing the capabilities of the latest Versatile Video Coding (VVC) standard.
To pioneer foundational research and accelerate the evolution of GFVC, this paper presents the first comprehensive survey of GFVC technologies, systematically bridging critical gaps between theoretical innovation and industrial standardization. In particular, we first review a broad range of existing GFVC methods with different feature representations and optimization strategies, and conduct a thorough benchmarking analysis. In addition, we construct a large-scale GFVC-compressed face video database with subjective Mean Opinion Scores (MOSs) based on human perception, aiming to identify the most appropriate quality metrics tailored to GFVC. Moreover, we summarize the GFVC standardization potentials with a unified high-level syntax and develop a low-complexity GFVC system which are both expected to push forward future practical deployments and applications. Finally, we envision the potential of GFVC in industrial applications and deliberate on the current challenges and future opportunities.
\end{abstract}

}

\maketitle

\IEEEdisplaynontitleabstractindextext

%
\IEEEpeerreviewmaketitle


%
%
%
%

 

\section{Introduction}

\IEEEPARstart{G}{enerative} Face Video Coding (GFVC)~\cite{chen2023generative,ultralow,9859867,CHEN2022DCC,icip2022zhao,konuko2023predictive,10811831,10743340} is a transformative technique in video compression by placing generative models at its core, redefining the tradeoff boundaries of compression efficiency and reconstruction quality.
Beyond traditional hybrid video coding frameworks~\cite{wiegand2003overview,sullivan2012overview,bross2021overview}, GFVC leverages the revolutionary advancements of deep generative models—such as Variational Auto-Encoders (VAEs)~\cite{VAE}, Generative Adversarial Networks (GANs)~\cite{goodfellow2014generative}, and Diffusion Models (DMs)~\cite{NEURIPS2021_49ad23d1}—to achieve unprecedented performance trade-offs between the compactness of coding bitstream and the quality of the compressed video. 
The key idea of GFVC is to encode the facial dynamics (\textit{e.g.,} landmarks, expressions and identity attributes)  into an ultra-compact latent representation at the encoder side and then employ a deep generative model to learn motion dependencies and reconstruct the face video at the decoder side.
This paradigm shift underscores generative models not merely as tools but as driving forces, enabling decoders to generate perceptual signal content from compact feature representations. 
By unifying compression with generation, GFVC exemplifies how generative AI is fundamentally reshaping the paradigm of video coding—prioritizing semantic compactness of coding bitstream and the enhanced perceptual quality of the reconstructed video.

\textbf{How to achieve the semantic compactness of coding bitstream?} In the past decades, generations of traditional video coding standards, such as H.264/Advanced Video Coding (AVC)~\cite{wiegand2003overview}, H.265/High Efficiency Video Coding (HEVC)~\cite{sullivan2012overview} and H.266/Versatile Video Coding (VVC)~\cite{bross2021overview}, have achieved promising Rate-Distortion (RD) performance and enabled a variety of broadcasting, streaming and conferencing applications. These widely adopted video coding frameworks with block-based motion estimation~\cite{jakubowski2013block} and transform coding~\cite{zhao2021transform} have been highly effective. However, they often struggle to preserve fine facial details and temporal dependencies in face videos. This limitation becomes particularly evident in low-bitrate scenarios, where the quality of the reconstructed video can degrade significantly, leading to artifacts such as blurring, blocking, and loss of facial structures and expressions.
Particular to talking-face video sequences, Model-Based Coding (MBC)~\cite{7268565,1457470,1989Object,lopez1995head,150969,364463} has been previously explored as an alternative paradigm to conventional coding frameworks.  MBC techniques, which could be dated back to the 1950s and rapidly developed in the 1990s, focus on encoding motion-based images or videos by using mathematical models and statistical representations tailored to the specific characteristics of the data. In particular, by leveraging these models (\textit{i.e.,} motion estimation models, or other statistical models), early MBC techniques sought to achieve high-efficiency compression while preserving essential visual information for reconstruction. However, limited by the not-so-powerful analysis-synthesis capabilities at that time, the RD performance has largely been unsatisfactory. 
Instead of directly encoding raw pixel values like traditional codecs, GFVC strictly follows the philosophy of MBC and tailors to the specific scenario in face video. As such, GFVC is capable of encoding face video into a compact latent representation and effectively reducing the amount of data required to represent the video, thus realizing more efficient storage and transmission.

\textbf{How to enhance the perceptual quality of the reconstructed video?} Motivated by the ongoing advances in deep learning techniques for various face-related tasks ~\cite{9868051,kong2025pixel, wang2018recurrent,10381809,10547422,9099607}, particularly in deep generative models~\cite{VAE,goodfellow2014generative,NEURIPS2021_49ad23d1} and face reenactment/animation models~\cite{FOMM,8954170,siarohin2021motion,10345691,10547422}, the concept of MBC has been effectively revitalized, leading to further development in GFVC. In particular, face image animation involves the process of creating lifelike facial expressions/movements in digital characters or avatars, where a series of poses or expressions at specific frames will be transferred to the reference frame such that the deep generative model can generate the animated video.
The connection between face video animation and MBC is significant, primarily in how both fields involve the representation and manipulation of facial data. This connection opens up opportunities for technology migration, especially for GFVC.
As such, the decoder of GFVC can be built upon a deep generative model like face animation model and the overall GFVC pipeline can follow the MBC's concept. Consequently, the reconstructed face video is not only visually realistic but also maintains fidelity to the original facial identity and expressions. 

By exploiting a strong facial prior for compact encoding and combining a powerful generative model for high-quality decoding, GFVC can realize promising compression efficiency for face video compression. Various efforts are made to improve representation compactness~\cite{chen2023csvt,ultralow,CHEN2022DCC}, reconstruction quality~\cite{konuko2022hybrid,icip2022zhao,konuko2023predictive}, modeling capacities~\cite{CHEN2025DCC,chen2025scalable}, reduced complexity~\cite{10743340,oquab2021low}, latency optimization~\cite{sivaraman2024gemino} and the diversity of applications~\cite{chen2023interactive}.
Despite substantial advancements, there remains a lack of a systematic survey and thorough analysis in the field of GFVC research. To this end, this paper aims to conduct a comprehensive survey of recent progress in GFVC techniques and standardization efforts, as well as provide a detailed benchmarking analysis of different algorithms and optimization strategies. The main contributions of this paper can be summarized as follows,

\begin{itemize}
\item{This paper provides a comprehensive overview of GFVC techniques, highlighting their key advancements/novelties and generalizing the common pipeline of these methodologies. Through a critical analysis of existing methodologies and benchmarking results, this paper seeks to inspire further research towards more efficient face video compression. } 
\item{This paper establishes a comprehensive GFVC-compressed face video database that incorporates subjective Mean Opinion Scores (MOSs) based on human perception. This database is designed to systematically evaluate the correlation between perceptual quality and varied objective assessment, with the aim of identifying the most suitable quality measurements for GFVC.}
\item{This paper outlines the standardization aspects of GFVC through the development of a unified high-level syntax and develops a low-complexity GFVC system by implementing lightweight model techniques. These efforts are expected to drive advancements in practical deployment and application scenarios. In addition, this paper provides a comprehensive analysis of application potentials, challenges and future research opportunities for the GFVC field.}
\end{itemize}

\begin{figure*}[tb]
\centering
\centerline{\includegraphics[width=1 \textwidth,height=7.5cm]{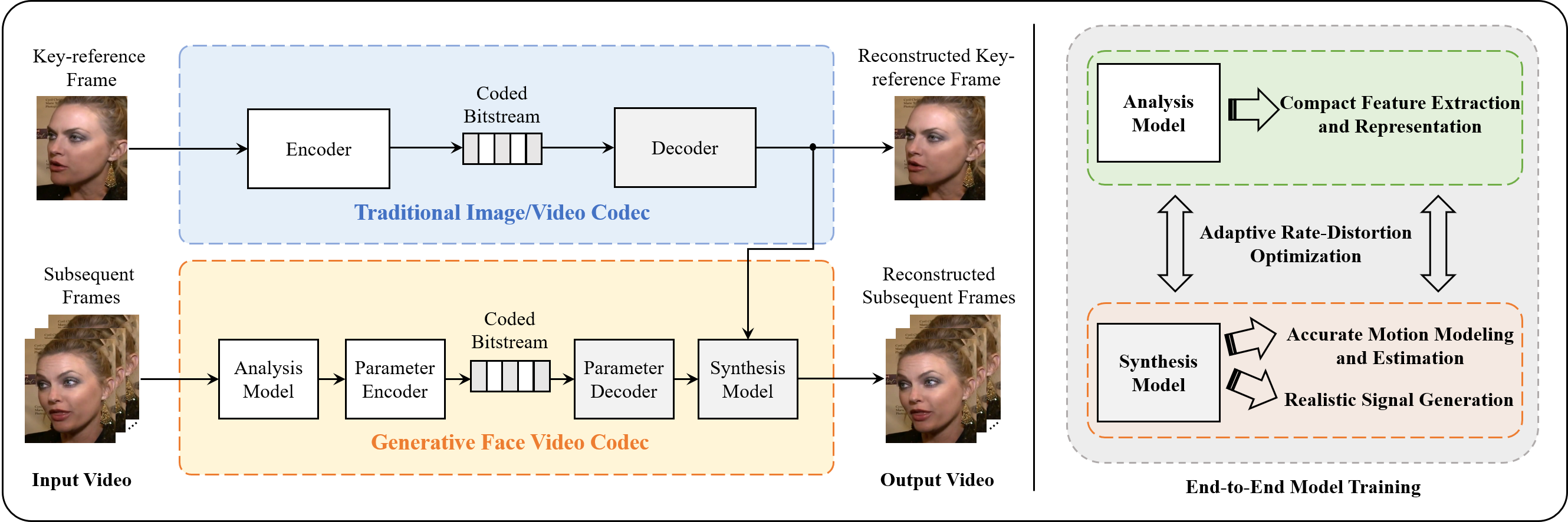}}  
\caption{Generalized pipeline and core techniques of Generative Face Video Coding~\cite{chen2023generative}.}
\label{fig1} 
\end{figure*}

\section{GFVC Formulation}
This section aims to formulate GFVC from the perspective of pipeline generalization and core techniques.

\subsection{Generalized Pipeline}
The goal of lossy video compression is to minimize perceived distortion (\textit{i.e.,} \(D\)) while adhering to a specified bitrate budget (\textit{i.e.,} \(R\)). This can also be framed as a rate-distortion optimization, where its objective is to reduce the total cost $\jmath _{cost}$ by balancing the trade-off between \(R\) and \(D\) as follows,
\begin{equation}
\begin{array}{c}
{
\jmath _{cost}=  D +\lambda \cdot R
},
\end{array}
\end{equation} where $\lambda$ is the Lagrange multiplier representing the \(R-D\) relationship for a particular quality level. 

However, in low bitrate environments, traditional coding methods may not effectively navigate this trade-off, resulting in either excessively low quality or inefficient compression. In particular, traditional coding methods typically rely on block-based motion estimation, transform coding and other strategies, which may not effectively balance this tradeoff for facial videos. In particular, traditional coding methods operate within these RD bounds based on Shannon's information theory, often relying on pixel-level processing and spatial-temporal redundancy reduction. However, facial videos contain highly complex and non-linear information that traditional methods struggle to represent efficiently. This inefficiency stems from their inability to exploit the underlying structure and semantics of facial data, leading to suboptimal compression rates. To conclude, traditional video coding frameworks struggle to optimize the RD trade-off for specialized content such as prior-based face video.


To overcome these above limitations of traditional hybrid video coding methods in face video compression, the emerging GFVC framework has been developed as illustrated in Fig. \ref{fig1}. The novel GFVC is built upon the philosophy of MBC and leverages the deep generative model to remedy MBC's weakness in poor analysis/synthesis abilities. In particular, the GFVC encoder primarily comprises three modules: an intra coder to compress the reference frame, an analysis model to capture the compact motion representation or facial structure of inter frames, and an encoding module to compress the compact motion representation into a decodable bitstream. Meanwhile, the GFVC decoder also consists of three key components: an intra decoder for reconstructing the reference frame, a parameter decoding module to reconstruct the compact representation, and a synthesis model responsible for face video reconstruction.

Assuming that $X_{t}$ represents the input face video clip at frame time t, where $X_{0}$ and $X_{1}^{t}$ are a intra frame (reference frame) and the subsequent inter frames, respectively.
First, the conventional image/video codec is employed to compress the intra frame $X_{0}$, aiming to provide the  image feature information for reconstructing subsequent frames. The reconstruction of the reference frame $\hat{X_{0}}$ can be represented as,
\begin{equation}
\hat{X_{0}}=Dec \left ( Enc\left ( X_{0} \right )  \right )  ,
\end{equation}  where $Enc\left ( \cdot  \right )$ and  $Dec\left ( \cdot  \right )$ represent the encoding and decoding processes of the traditional intra coding. 

Subsequently, the inter frames $X_{1}^{t}$ are input into an analysis model to extract the compact motion parameters $\theta_{1}^{t}$. The extraction process $\mathcal{E} \left (\cdot \right )$  can be described as,
\begin{equation}
\theta_{1}^{t} = \mathcal{E} \left ( X_{1}^{t}\right ) ,
\end{equation} where this learning-based compact feature representation process can decrease the amount of coding bits required and enhance the compression of facial videos at significantly lower bitrates. Afterwards, the extracted compact representations $\theta_{1}^{t}$ are efficiently inter predicted, quantized, and entropy coded into a bitstream using the parameter encoding module.

Upon receiving the bitstream of compact representations, the decoder undertakes the relevant parameter decoding steps (like inverse quantization and compensation) to reconstruct the compact representation $\hat{\theta_{1}^{t}}$. Finally, the reconstruction of the facial video is accomplished by utilizing the reconstructed compact representation $\hat{\theta_{1}^{t}}$ and the decoded reference frame $\hat{X_{0}}$ through the frame synthesis model. Specifically, $\hat{X_{0}}$ is further transformed into the compact representation $\theta_{0}$ via the analysis model, enabling the generation of motion fields between $\theta_{0}$ and $\hat{\theta_{1}^{t}}$. Consequently, the frame synthesis model leverages the robust inferential capabilities of deep generative networks to reconstruct high-quality video frames $\hat{X_{1}^{t}}$ under the guidance of $\hat{X_{0}}$. These processes can be outlined as,
\begin{equation}
\hat{X_{1}^{t}} =\zeta \left ( \hat{X_{0}} ,\varpi \left ( \hat{\theta_{1}^{t}} , \mathcal{E} \left ( \hat{X_{0}} \right ) \right )  \right ) ,
\end{equation} where $\varpi \left ( \cdot  \right )$ and  $\zeta\left ( \cdot  \right )$ represents the motion filed calculation and frame generation processes, respectively.

Generally, the overall loss is introduced to optimize the GFVC model training, such that the optimal performance can be achieved. It can be formulated as, 
\begin{equation}
\begin{array}{c}
{
\mathop{\arg\min} \sum  \mathcal{L}\left ( X_{1}^{t} ,\hat{X_{1}^{t}}\right ) + \lambda \cdot R(\theta_{1}^{t})
},
\end{array}
\end{equation} where $\mathcal{L}\left ( \cdot\right )$ represents the function to compute the loss between the generated face image and the ground truth $X_{1}^{t}$, $\hat{X_{1}^{t}}$, $\mathcal{R}\left ( \cdot\right )$ denotes the rate consumption of the compression parameters, and $\mathop{\arg\min}$ optimizes the network parameters to minimize the overall loss across all frames. It should be mentioned that the end-to-end training strategy is employed to optimize the entire GFVC system, including components like the feature extraction model, deep generative model, and other modules, simultaneously allowing all components to learn jointly and adapt to each other during training. As such, the GFVC model can learn to encode, generate, and compress face video sequences more effectively, leading to improved performance and potentially better generalization on unseen data (refers to data that the machine learning model has not been exposed to during training).

\subsection{Core Techniques}
The core of GFVC lies in deep feature extraction, generative model-based reconstruction, and dynamic modeling, facilitating remarkable compression performance in face videos. The core techniques and operational principles of GFVC can be summarized as follows,
\begin{itemize}
\item{\textbf{Compact Feature Representation and Extraction:} GFVC initially extracts crucial features like facial contours, expressions, and poses from input facial videos, where these features are typically learned using deep neural networks and represented with low-dimensional representation. Unlike traditional video coding methods that process video data at the pixel level, GFVC focuses on representing the video in terms of these high-level features, which are more compact and efficient for encoding.} 
\item{\textbf{Accurate Motion Modeling and Estimation:} Facial videos often contain dynamic changes, such as facial expressions and head movements, which are challenging for traditional video coding techniques to handle efficiently. GFVC addresses this by incorporating motion estimation and expression modeling techniques. For example, optical flow estimation or 3D facial models can be used to capture these dynamic changes. The motion and expression information is encoded alongside the feature vectors and used at the decoder to guide the generative model to accurately reconstruct the video frames.} 
\item{\textbf{Realistic Signal Generation:} The core of GFVC is a generative model, often based on architectures like GANs or VAEs. The generative model takes the extracted compact feature representations as input and reconstructs the facial video frames at the decoder. Instead of transmitting raw pixel data, GFVC transmits these compact feature representation, which are then used by the generative model to transform dense motion fields and reconstruct the video frames. This approach drastically reduces the amount of data that needs to be transmitted while maintaining high visual quality in the reconstructed video.} 
\item{\textbf{Adaptive Rate-Distortion Optimization:} GFVC employs adaptive optimization strategies to dynamically adjust encoding parameters or refresh reference frame based on the video content. For instance, it can allocate different bitrates for different motion complexity using the scalable representation or progressive tokens. This ensures an optimal balance between compression efficiency and visual quality, and achieves bandwidth intelligence for GFVC. } 
\item{\textbf{End-to-End Model Training:} GFVC is typically trained in an end-to-end manner, making the joint optimization of the entire pipeline (\textit{i.e.,} feature extraction, motion estimation, and frame generation) feasible. Training is performed on large-scale datasets of facial videos, allowing the system to learn the underlying distribution of facial features and dynamics. This end-to-end training ensures that the system can efficiently compress and reconstruct diverse facial videos with high fidelity. } 
\end{itemize}

\section{Overview of Progress in Recent Years}
\subsection{Literature Overview}

\begin{table*}[!t]
\vspace{-2.5em}
\renewcommand\arraystretch{1.35}
\caption{Summary of GFVC methods with different compact feature representations and optimization strategies}  
\label{table_summary_research}
\centering
\resizebox{1\textwidth}{!}{
\begin{tabular}{lllm{10.5cm}} 
\hline
\textbf{Category}                                                                               & \textbf{Method Name} & \textbf{Published Time} & \textbf{Description}                                                                                                                                                                                            \\ \hline
\multirow{12}{*}{\begin{tabular}[c]{@{}l@{}}Compact \\ Representations\end{tabular}} & FOMM~\cite{FOMM}                 & NeurIPS 2019            & Represents facial motion using 2D unsupervised keypoints and its affine transformation. Although not being a compression technique itself, FOMM could be regarded as the basis of keypoint-based GFVC algorithms. \\
                                                                                                & DAC~\cite{ultralow}                  & ICASSP 2021             & Employs FOMM as motion representation that could compress the unsupervised 2D keypoints to achieve promising compression efficiency, and adds an Intra-Refresh mechanism to reduce error drift and achieve rate variability.                                                                                        \\
                                                                                                & VSBNet~\cite{9455985}               & ICMEW 2021              & Exploits 2D landmarks to develop   a Visual-Sensitivity-Based network, aiming to improve face fidelity for ultra-low bitrate communication.                                                                                     \\
                                                                                                & FV2V~\cite{wang2021Nvidia}         & CVPR 2021               & Disentangles the complex face   signal into 3D keypoints and its translation/rotation matrices, which  can achieve free-view control and ultra-low bitrate face video communication.                                            \\
                                                                                                & SNRVC~\cite{9810784}                & DCC 2022                & Characterizes face signal with a   series of compact facial semantics for reconstructing face video under the low-bitrate constraints.                                                                                          \\
                                                                                                & CFTE~\cite{CHEN2022DCC}                 & DCC 2022                & Proposes to characterize the   temporal evolution of face video with compact feature representation based upon an end-to-end framework.                                                                                         \\
                                                                                                & SOM~\cite{9949138}                 & MMSP 2022                & Employs the second order motion coherency to improve the temporal smoothness of the FOMM model. 
                                                                                                \\
                                                                                                & TPS~\cite{9880299}                  & CVPR 2022               & Develops a thin-plate spline   motion estimation scheme that can produce a more flexible optical flow for high-quality reconstruction.                                                                                          \\
                                                                                                & MTTF~\cite{yin2024generative}                 & arXiv 2024              & Proposes a motion factorization  strategy to implicitly characterize multi-granularity feature representations.                                                         \\
                                                                                                &  FVC-3K2M~\cite{10811831}                 & TIP 2024                & Proposes a 3D-Keypoint-and-2D-Motion based generative method for Face Video Compression that well ensure perceptual compensation and visual consistency between motion description and face reconstruction.                                                                                            \\                                                                                                 
                                                                                                & Bi-LFAC~\cite{riku2025DCC}              & DCC 2025                & Develops a bidirectional learned   facial animation codec for ultra-low bitrate face video communication.                                                                                                                         \\
                                                                                                & AT-GFVC~\cite{Lin2025DCC}              & DCC 2025                & Designs an affine   transformation-based compression framework to handle large motion scene.                                                                                                                                      \\                                                                                            
                                                                                                & PFVC~\cite{CHEN2025DCC}                 & DCC 2025                & Utilizes adaptive visual tokens to realize exceptional trade-offs between reconstruction robustness and bandwidth intelligence.                                                                                               \\ 
                                                                                                                                                                                                & IFVC~\cite{chen2023generative}                 & TIP 2025              & Develops a novel interactive coding framework that can represent face signal with a series of facial semantics and allow humans to interact with these representations instead of   the raw signals.                          \\                                                                                                
                                                                                                \hline
\multirow{16}{*}{\begin{tabular}[c]{@{}l@{}}Optimization \\ Strategies\end{tabular}}            & Mob M-SPADE~\cite{oquab2021low}          & CVPRW 2021              & Employs the SPADE architecture   and segmentation maps to develop the first real-time GFVC system on the mobile CPU.                                                                                                            \\
                                                                                                & MAX-RS~\cite{volokitin2022neural}               & CVPRW 2022              & Explores multiple source frames for multi-view neural face video compression via the view aggregation and selection techniques.                                                                                               \\
                                                                                                & DMRGP~\cite{icip2022zhao}                & ICIP 2022               & Designs a dynamic   multi-reference prediction mechanism for GFVC, such that GFVC can well handle large head motion scene.                                                                                                      \\
                                                                                                & HDAC~\cite{konuko2022hybrid}                 & ICIP 2022               & Proposes a layered coding scheme   that can facilitate GFVC to improve long-term dependencies and alleviate background occlusions.                                                                                              \\
                                                                                                & FOMM\_BE~\cite{9949509}                 & MMSP 2022               & Proposes a background enhancement method to improve user experiences for FOMM-based video reconstruction algorithms. \\
                                                                                                & CVC STR~\cite{compressing2022bmvc}              & BMVC 2022               & Employs a frame interpolator for the reduction in the temporal redundancy and a patch-wise super-resolution   algorithm for the reconstruction quality improvement.                                                             \\
                                                                                                & Bi-Net~\cite{9859867}               & ICME 2022               & Develops a bitrate-adjustable   hybrid GFVC framework that is integrated with pixel-wise bi-prediction, low-bitrate-FOMM algorithm and lossless compression.                                                                    \\
                                                                                                & CTTR~\cite{chen2023csvt}                 & TCSVT 2023              & Employs the spatial-temporal   adversarial training to ensure temporal consistency  and develops a dynamic reference refresh   scheme to improve the robustness against large head-pose motions.                                  \\
                                                                                                & RDAC~\cite{konuko2023predictive}                 & ICIP 2023               & Proposes a predictive coding   scheme that can additionally compress the residual information to improve the   reconstruction quality of the GFVC model.                                                                          \\
                                                                                                & GFVC\_Translator~\cite{yin2024parametertranslator}     & DCC 2024                & Develops a face feature transcoding scheme than enable GFVC's interoperability across different   encoders and decoders.                                                                                                        \\
                                                                                                & HDAC+~\cite{10772980}                & EUVIP 2024              & Exploits a dual residual   learning strategy for improved GFVC performance.                                                                                                                                                       \\
                                                                                                & MRDAC~\cite{MRDAC}                & MMSP 2024               & Propose the multiple reference animation and contrastive learning formulation to minimize GFVC reconstruction drift, especially when used in a bi-directional   prediction mode.                                               \\
                                                                                                & Lite-CFTE~\cite{10743340}            & MMSP 2024               & Develops   a practical lightweight strategy for the CFTE model, aiming to provide insights into practical deployments and efficient inference.                                                                                  \\
                                                                                                & MultiRes GFVC~\cite{10743918}        & MMSP 2024               & Focuses on extending support for multi-resolution reconstruction and reducing model complexity.                                                                                                                               \\
                                                                                                & HDAC-HF~\cite{10849880}              & VCIP 2024               & Introduces a high-frequency shuttling mechanism to enhance reconstruction fidelity.                                                                                                                                             \\
                                                                                                & Gemino~\cite{105555}               & NSDI 2024               & Proposes a neural compression   system for video conferencing based on high-frequency conditional super resolution  and implements this system via WebRTC for real-time inference.                                           \\
                                                                                                & PGen~\cite{chen2025scalable}                 & arXiv 2025              & Leverages scalable   representation and layered reconstruction for GFVC, aiming at bandwidth intelligence and reconstruction robustness.                                                                                        \\ \hline
\end{tabular}
}
\end{table*}

\begin{figure}[tb]
\centering
\centerline{\includegraphics[width=0.48\textwidth]{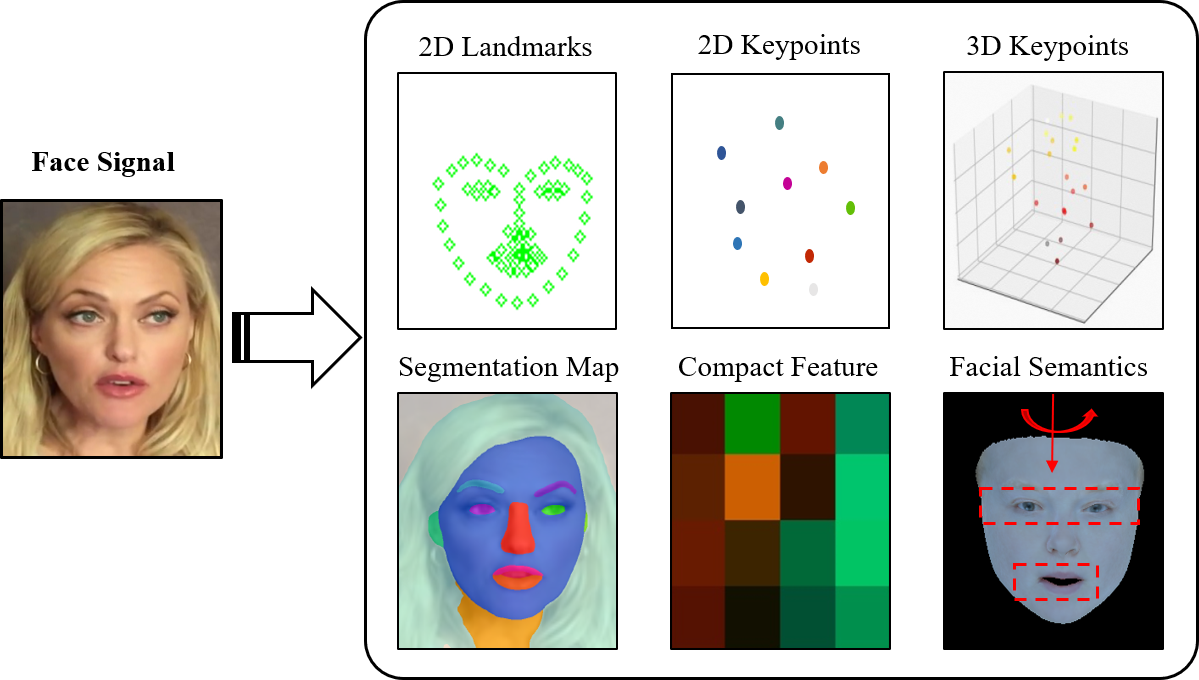}}  
\caption{Illustrations of compact representations for face signal.}
\label{compact_representation} 
\end{figure}

As illustrated in TABLE \ref{table_summary_research}, we review the progress of GFVC in recent years and categorize these advanced methods with two types: compact visual representations and optimization strategies.

\subsubsection{Compact Representations}

Compact visual representations are pivotal for GFVC as they enable efficient encoding of facial dynamics while minimizing data redundancy. By characterizing complex facial motions (\textit{e.g.,} expressions and head movements) with low-dimensional features like 2D keypoints or latent vectors, these representations can reduce bandwidth requirements and computational overhead. This is critical for real-time applications like teleconferencing, where high compression ratios must coexist with perceptual fidelity. Additionally, compact representations can enhance robustness in low-bandwidth scenarios by prioritizing semantically meaningful facial attributes over raw pixel data.

As shown in Fig. \ref{compact_representation}, there has been a growing interest in exploring novel techniques for generating compact visual representations in the context of face video compression. In particular, some studies have focused on leveraging face models (\textit{e.g.,} 2D landmark detection model, segmentation model or 3D morphable model) for facial feature extraction and representation learning, which enables the generation of compact yet informative visual representations suitable for compression algorithms. These compact visual representations include but are not limited to: 2D landmarks~\cite{9455985}, segmentation map~\cite{oquab2021low} and 3D facial semantics~\cite{9810784,chen2023interactive}. Researchers have also investigated the representations of facial keypoints~\cite{FOMM,ultralow}, compact feature~\cite{chen2023csvt,CHEN2022DCC}, motion vectors~\cite{yin2024generative}, and other low-dimensional descriptors~\cite{9880299} to encode facial dynamics effectively. By capturing the essence of facial expressions and movements in an unsupervised manner, these representations not only facilitate efficient compression, but also enable high-quality reconstruction of facial videos with minimal loss of information.

Overall, the development of compact visual representations in GFVC tasks is a vibrant research area that holds significant potential for enhancing video compression efficiency, enabling real-time applications, and improving the visual quality of compressed face videos.

\subsubsection{Optimization Strategies}

In addition to compact representations, there are a series of optimization strategies employed to enhance GFVC performance:

\begin{itemize}
\item{\textbf{Dynamic Reference~\cite{chen2023csvt,icip2022zhao,MRDAC,volokitin2022neural}:} Dynamically selecting reference frames based on complex motion changes improves compression efficiency and reduces artifacts in GFVC applications. It helps adapt to varying facial movements and ensures high-quality reconstructions in dynamic scenes.} 
\item{\textbf{Bidirectional Prediction~\cite{9859867,MRDAC}:} Utilizing both past and future frames for prediction enhances motion estimation accuracy, especially for complex facial dynamics. This strategy reduces prediction errors, lowers bitrate requirements, and improves temporal coherence in reconstructed videos.} 
\item{\textbf{Residual Coding~\cite{konuko2023predictive,10772980}:} Encoding residual differences between GFVC-generated and actual frames could facilitate to preserve fine details critical for perceptual realism. This approach balances compression efficiency with visual fidelity, ensuring that subtle facial features are accurately reconstructed.} 
\item{\textbf{Scalable/Layered Coding~\cite{chen2025scalable,konuko2022hybrid,10849880}:}  Implementing scalable or layered coding techniques allows for adaptive video streaming based on network conditions and device capabilities. This strategy optimizes bandwidth usage and ensures smooth video delivery across various platforms.} 
\item{\textbf{Model Interoperability~\cite{yin2024parametertranslator}:} Ensuring compatibility and seamless integration between different GFVC models enhances the overall performance and flexibility of communication systems. It enables efficient information exchange and collaboration between diverse model architectures.} 
\item{\textbf{Model Lightweight~\cite{oquab2021low,10743340}:} Developing lightweight models tailored for GFVC tasks optimizes computational resources while maintaining high-quality compression. Lightweight models are essential for real-time applications and devices with limited computational resources.} 
\item{\textbf{Multi-Resolution Support~\cite{10743918}:} Supporting multiple resolutions allows for adaptive streaming and efficient utilization of network bandwidth. This strategy ensures that videos can be delivered at varying quality levels based on device capabilities and network conditions.}  
\item{\textbf{Super-Resolution/Enhancement Operations~\cite{compressing2022bmvc}:} Implementing super-resolution or enhancement techniques enhances the visual quality of reconstructed videos by upscaling low-resolution content to higher resolutions. This operation improves facial detail preservation and overall fidelity in GFVC tasks.}  
\item{\textbf{WebRTC Communication~\cite{105555}:} Utilizing WebRTC communication protocols facilitates real-time video streaming and communication over the web. This strategy enhances the efficiency of GFVC tasks by enabling low-latency, high-quality video transmission in web-based applications.} 
\end{itemize}

These optimization strategies collectively contribute to improving compression efficiency, reducing artifacts, preserving facial details, and achieving high-fidelity reconstructions in GFVC tasks. By integrating these techniques, the performance and practical applicability of GFVC systems can be greatly boosted.

\subsection{Quantitative Benchmark Results}
\label{benchmark_result}
To compare the RD performance of GFVC techniques with different feature representations and optimization strategies, a quantitative benchmark is built by evaluating several selected representative GFVC algorithms under the same training and testing conditions.

\subsubsection{Experimental Settings}
\label{sec:CTC} 

\quad \textbf{a}) \textit{\textbf{Datasets}}:\quad For training of benchmark models, a large-scale face video dataset constructed with 5,000 videos from VoxCeleb2~\cite{2018VoxCeleb2} dataset and 5,000 videos from CelebV-HQ~\cite{zhu2022celebvhq} dataset are selected, yielding totally 10,000 videos with the resolution of 512$\times$512. Specifically, 
\begin{itemize}
\item VoxCeleb2~\cite{2018VoxCeleb2} is an audio-visual dataset consisting of short clips of human speech, extracted from interview videos uploaded to YouTube. All VoxCeleb2 data mainly focuses on face-centric content.
\item CelebV-HQ~\cite{zhu2022celebvhq} is a large-scale, high-quality, and diverse video dataset with diverse facial attribute annotations and head-and-shoulder content.
\end{itemize}

For evaluation of benchmark models, we follow the GFVC standardization activities in Joint Video Experts Team (JVET) of ISO/IEC JTC 1/SC 29 and ITU-T SG21 (formerly SG16) and adopt the test sequences specified in JVET-AJ2035~\cite{JVET-AJ2035}. Specifically, there are four classes of sequences in the test set,
\begin{itemize}
\item Class A contains 15 sequences with face-centric contents, and each sequence has 250 frames with 256$\times$256 resolution.
\item Class B contains 18 sequences with head-and-shoulder contents, and each sequence has 125 frames with 256$\times$256 resolution.
\item Class C contains 15 sequences with face-centric contents, and each sequence has 250 frames with 512$\times$512 resolution.
\item Class D contains 18 sequences with head-and-shoulder contents, and each sequence has 125 frames with 512$\times$512 resolution.
\end{itemize}

\textbf{b}) \textit{\textbf{Implementation Details}}:\quad We implement all benchmark models with the Pytorch framework and use NVIDIA TESLA A100 GPUs for training. In particular, these models are trained with 50 epochs via the Adam optimizer with $\beta_{1} = 0.5$, $\beta_{2}=0.999$. 
Besides, we set the initial learning rate as $0.0002$ and use multi-step learning rate scheduler with $\gamma=0.1$ and  $milestone=[30,45]$.

\textbf{c}) \textit{\textbf{Quality Evaluation Metrics}}:\quad For objective quality evaluation, we follow the common practice of generative face video coding and its standardization effort~\cite{chen2023generative,JVET-AJ2035} and choose two perceptual-level metrics(~\textit{i.e.,} Learned Perceptual Image Patch Similarity~(LPIPS)~\cite{lpips} and Deep Image Structure and Texture Similarity~(DISTS)~\cite{dists}) as well as two pixel-level metrics (~\textit{i.e.,} Peak Signal Noise Ratio~(PSNR) and Structure Similarity~(SSIM)~\cite{2004Image}). All metrics are measured in RGB444 domain. Bjøntegaard-delta-rate~(BD-rate)~\cite{Bjntegaard2001CalculationOA} and RD curve are adopted to quantify the overall compression performances between the GFVC methods and the VVC anchor. To display the RD curves in an increasing manner and calculate BD-rate saving, we use ``1-DISTS'', ``1-LPIPS'' as the y-axis of the graphs.

\subsubsection{Selected Comparison Algorithms}

\quad \textbf{a}) \textit{\textbf{VVC Anchor}}:\quad Following the test procedure in~\cite{JVET-AJ2035}, we adopt the Low-Delay-Bidirectional (LDB) configuration in VTM 22.2 reference software for VVC~\cite{bross2021overview}, where the quantization parameters (QP) are set to 37, 42, 47 and 52. To adapt to extended bit-rate coverage of further optimized algorithms, the rate point of QP 32 is added for comparison of higher bit-rates. 

\textbf{b}) \textit{\textbf{GFVC with Different Feature Representations}}:\quad we choose five algorithms for comparison, \textit{i.e.,} FOMM~\cite{FOMM}, DAC~\cite{ultralow}, CFTE~\cite{CHEN2022DCC}, FV2V~\cite{wang2021Nvidia} and TPS~\cite{9880299}. The details are listed as follows,
\begin{itemize}
\item FOMM~\cite{FOMM} uses ten 2D key-points and their corresponding Jacobian matrices as face parameters. Each key-point has two coordinates and four matrices' elements, yielding total 60 parameters for each inter-frame.
\item DAC~\cite{ultralow} utilizes ten 2D key-points as face parameters. Each key-point has two coordinates, yielding total 20 parameters for each inter-frame.
\item CFTE~\cite{CHEN2022DCC} leverages 4$\times$4 compact matrix to represent the temporal evolution, yielding total 16 parameters for each inter-frame.
\item FV2V~\cite{wang2021Nvidia} uses 15 3D key-poitns as well three expression parameters and a 3 $\times$ 3 rotation matrix to enable the free-view control of talking-head, yielding total 57 parameters.
\item TPS~\cite{9880299} utilizes 10 thin-plate spline~(TPS) transforms to characterize the motion deformation, and each TPS transform is estimated by a five-key-point-pair, yielding total 100 parameters.

\end{itemize}
For these five basic GFVC algorithms, they are implemented as multi-resolution models~\cite{JVET-AJ0052} to support both 256$\times$256 and 512$\times$512 resolutions. For performance evaluation, we follow the general pipeline in Fig.~\ref{fig1} and test procedure in~\cite{JVET-AJ2035}, where the intra mode of the VTM 22.2 software with the QPs of \{22, 32, 42, 52\}, and inter frame features are compressed via a context-adaptive arithmetic coder. The corresponding GFVC performances are compared with the VVC QP point of \{37, 42, 47, 52\}.

\textbf{c}) \textit{\textbf{GFVC with Different Optimization Strategies}}:\quad we choose three representative optimization strategies and their corresponding models, \textit{i.e.,} HDAC~\cite{konuko2022hybrid}, RDAC~\cite{konuko2023predictive} and MRDAC~\cite{MRDAC}. These three algorithms use DAC~\cite{ultralow} as basic animation-based model, and aim to improve the prediction quality and bit-rate coverage with layered coding, residual coding and bidirectional dynamic reference techniques, respectively. The details are listed as follows,
\begin{itemize}
\item HDAC~\cite{konuko2022hybrid} adopts a multi-layer structure, where VVC codec is regarded as base layer and a DAC-like codec is utilized as GFV layer. At the decoder, the animation-based generation results from the two layers are fused to obtain the final reconstruction. In this benchmark, we set the VVC base layer as a low quality layer with QP 42, and GFV layer as a high quality layer with key frame QP \{12, 22, 32, 42\}.

\item RDAC~\cite{konuko2023predictive} utilizes learned image codec to compress the residual signals between original inter frames and their reconstructions by DAC-like codec. At the decoder side, the residual signals are fused to animation-based generation to obtain final reconstruction. In this benchmark, we set the key frame QP of DAC as \{17, 22, 27, 32\}. And to avoid error accumulation, the residual calculation is renewed with a window size of 8.

\item MRDAC~\cite{MRDAC} proposes to use multiple key frames fusion with contrastive learning to improve the prediction accuracy of final reconstruction. Its key frames are compressed by VVC codec with the same quality and improving the final reconstruction by increasing the number of key frames. In our implementation, we set key frame QP as 22 and the number of key frames as \{1, 2, 3, 4\}.

\end{itemize}
For these three extended GFVC algorithms, they are implemented on 256$\times$256 resolution. For BD-Rate saving against VVC anchor, they are compared to rate point of QP \{32, 37, 42, 47\}.

\subsubsection{Objective Performances}
\begin{figure}[t]
\centering
\subfloat[256$\times$256: Rate-DISTS]{\includegraphics[width=0.248\textwidth]{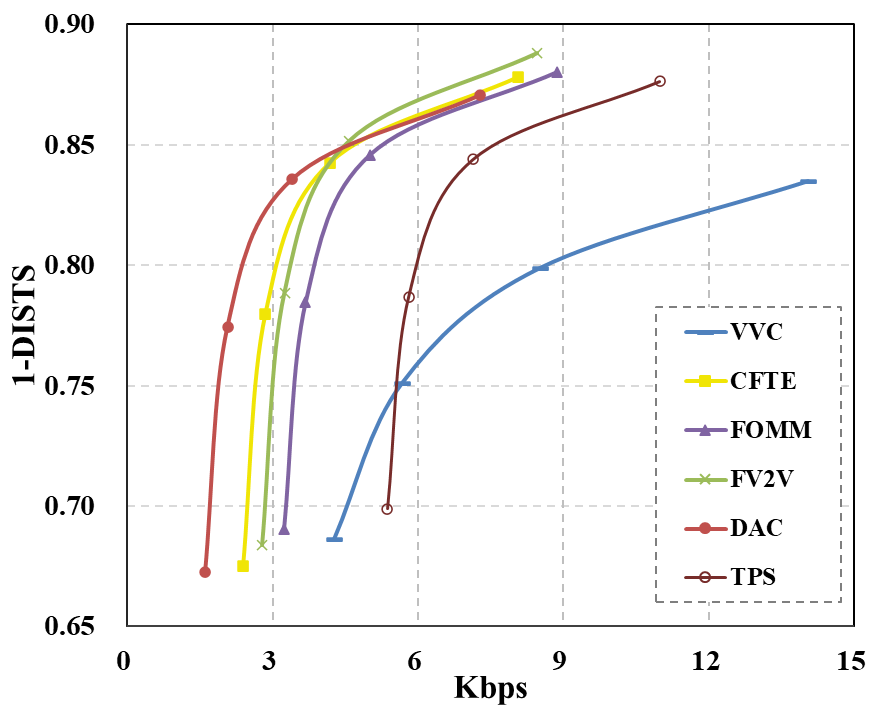}}
\subfloat[256$\times$256: Rate-LPIPS]
{\includegraphics[width=0.248\textwidth]{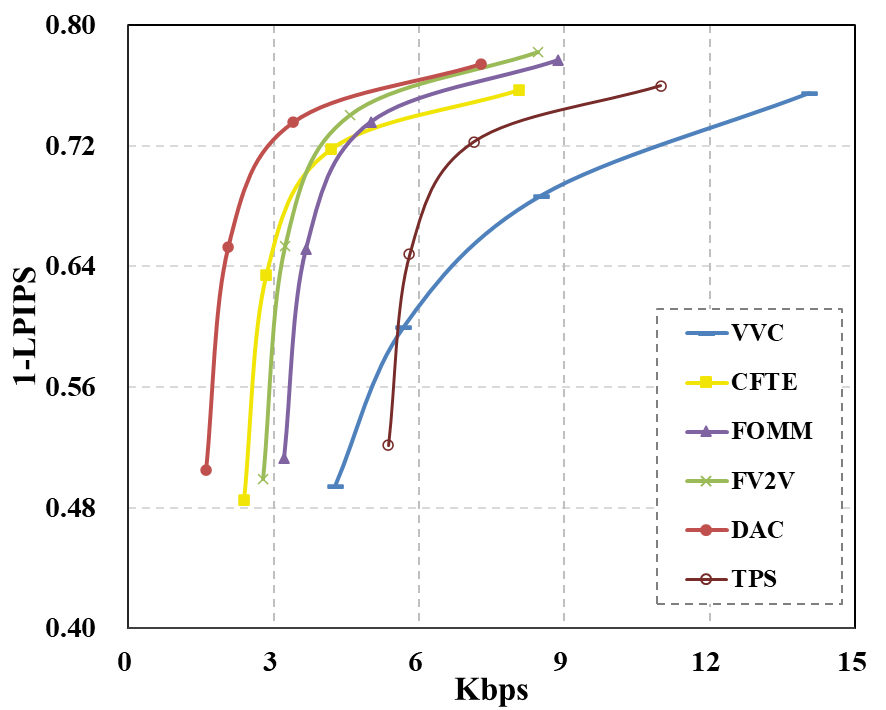}}
\\
\subfloat[512$\times$512: Rate-DISTS]{\includegraphics[width=0.248\textwidth]{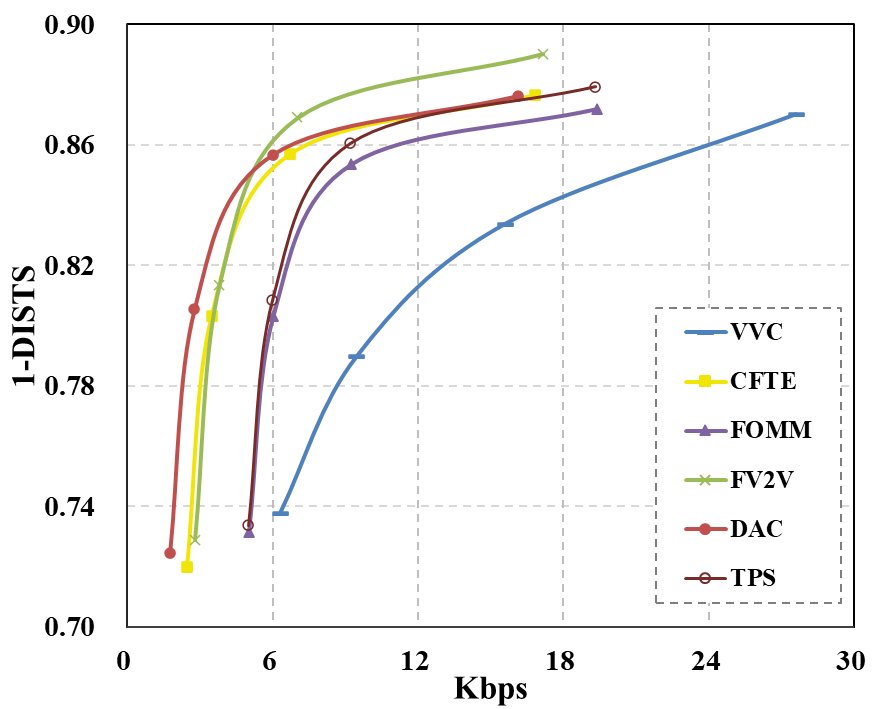}}
\subfloat[512$\times$512: Rate-LPIPS]
{\includegraphics[width=0.248\textwidth]{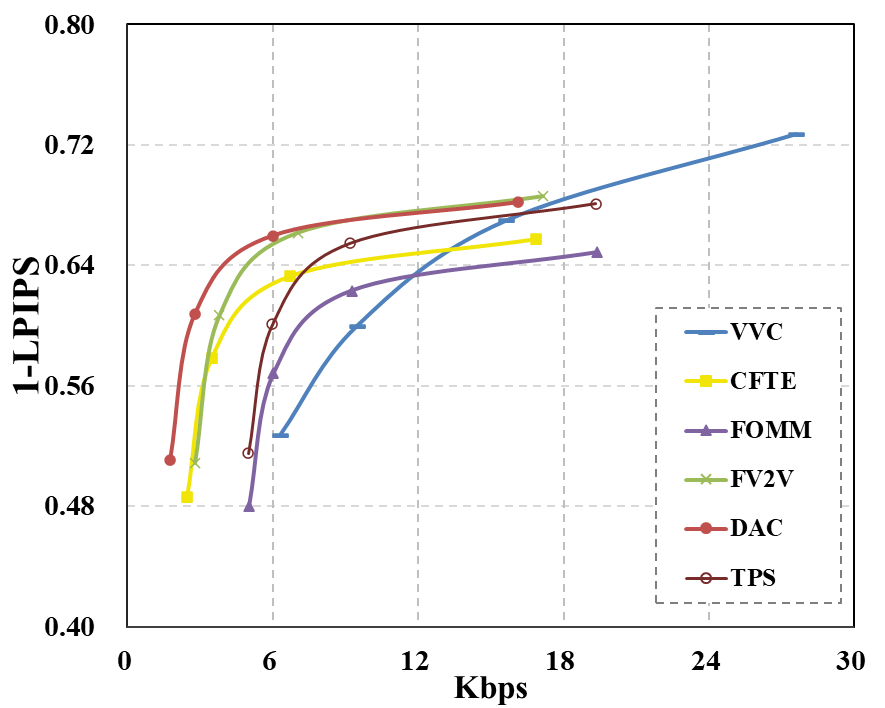}}
\caption{Rate-Distortion performance comparisons of VVC~\cite{bross2021overview} and five basic GFVC benchmark (FOMM~\cite{FOMM}, CFTE~\cite{CHEN2022DCC}, FV2V~\cite{wang2021Nvidia}, DAC~\cite{ultralow} and TPS~\cite{8099774}) in terms of rate-DISTS and rate-LPIPS for both 256$\times$256 and 512$\times$512 resolutions.} %
\label{benchmark_main_result_perceptual}
\end{figure}

\begin{table}[t]
\caption{RD performance comparisons on five basic GFVC algorithms in terms of average BD-rate savings over the VVC~\cite{bross2021overview} anchor on 256$\times$256 resolution}
    \label{bd_main_256}
    \centering
\renewcommand\arraystretch{1.1}
\begin{tabular}{ccc}
\hline
Algorithm      & Rate-DISTS & Rate-LPIPS \\ \hline
FOMM~\cite{FOMM}         & -44.71\%            & -41.10\%                      \\
CFTE~\cite{CHEN2022DCC}           & -55.56\%            & -49.39\%                      \\
FV2V~\cite{wang2021Nvidia}           & -52.33\%            & -46.51\%                      \\
DAC~\cite{ultralow}            & -66.99\%            & -65.73\%                      \\
TPS~\cite{9880299}  & -15.37\%            & -11.40\%                     \\
 \hline
\end{tabular}
\vspace{-0.2cm}
\end{table}

\begin{table}[t]
\caption{RD performance comparisons on five basic GFVC algorithms in terms of average BD-rate savings over the VVC~\cite{bross2021overview} anchor on 512$\times$512 resolution}
    \label{bd_main_512}
    \centering
\renewcommand\arraystretch{1.1}
\begin{tabular}{ccc}
\hline
Algorithm      & Rate-DISTS & Rate-LPIPS \\ \hline
FOMM~\cite{FOMM}         & -39.27\%            & -12.36\%                      \\
CFTE~\cite{CHEN2022DCC}           & -64.00\%            & -47.34\%                      \\
FV2V~\cite{wang2021Nvidia}           & -66.16\%            & -52.67\%                      \\
DAC~\cite{ultralow}            & -70.56\%            & -65.77\%                      \\
TPS~\cite{9880299}  & -41.07\%            & -28.54\%                     \\
 \hline
\end{tabular}
\vspace{-0.2cm}
\end{table}

\begin{figure}[t]
\centering
\subfloat[256$\times$256: Rate-PSNR]{\includegraphics[width=0.248\textwidth]{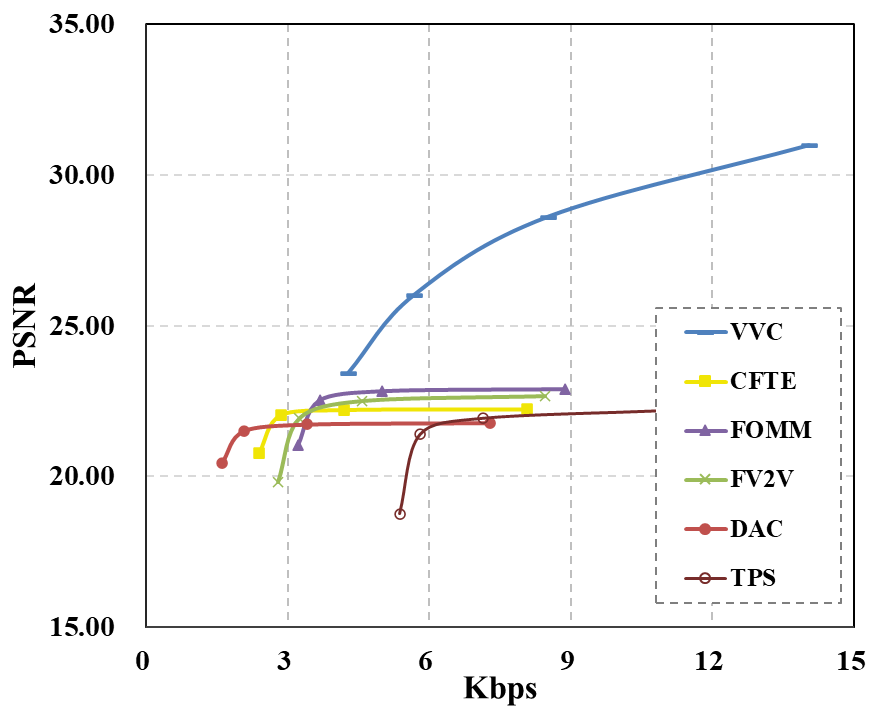}}
\subfloat[256$\times$256: Rate-SSIM]
{\includegraphics[width=0.248\textwidth]{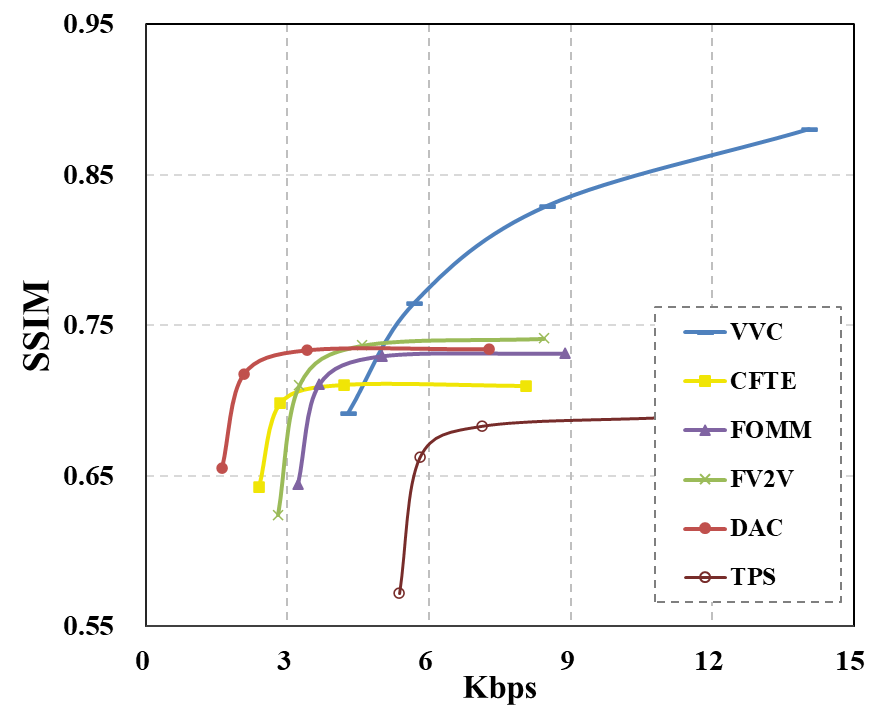}}
\\
\subfloat[512$\times$512: Rate-PSNR]{\includegraphics[width=0.248\textwidth]{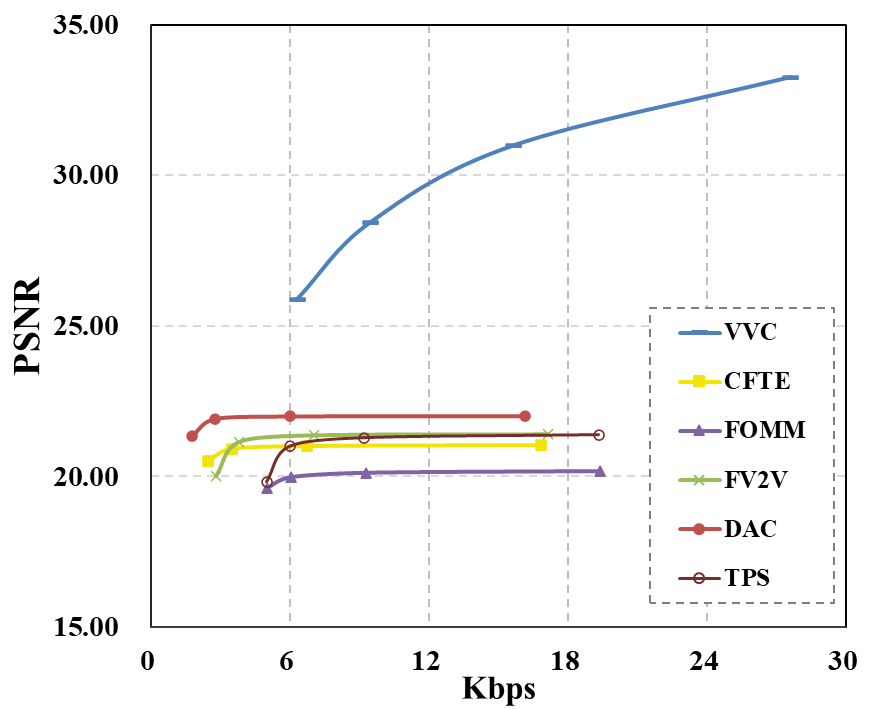}}
\subfloat[512$\times$512: Rate-SSIM]
{\includegraphics[width=0.248\textwidth]{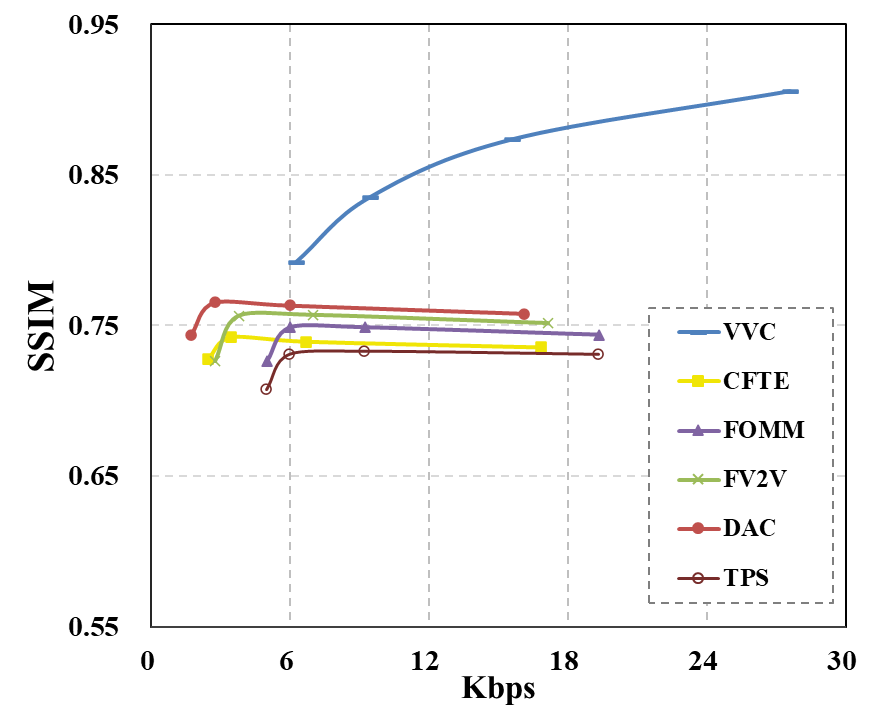}}
\caption{Rate-Distortion performance comparisons of VVC~\cite{bross2021overview} and five basic GFVC benchmark (FOMM~\cite{FOMM}, CFTE~\cite{CHEN2022DCC}, FV2V~\cite{wang2021Nvidia}, DAC~\cite{ultralow} and TPS~\cite{8099774}) in terms of rate-PSNR and rate-SSIM for both 256$\times$256 and 512$\times$512 resolutions.} %
\label{benchmark_main_result_pixel}
\end{figure}

\quad \textbf{a}) \textit{\textbf{Basic GFVC Algorithms}}:\quad The RD performance of five basic GFVC algorithms in terms of two perceptual-level qualities are shown in Fig.~\ref{benchmark_main_result_perceptual}. It can be observed that, most of GFVC algorithms can outperform VVC with a large margin, illustrating the effectiveness of GFVC pipeline for talking-face communication under ultra-low bit-rate.
Specifically, DAC~\cite{ultralow} achieves the lowest bit-rate and FV2V~\cite{wang2021Nvidia} achieves the best reconstruction quality. TPS~\cite{9880299} shows the highest bit-rate consumption because of its larger scale feature representation. TABLE~\ref{bd_main_256} and TABLE~\ref{bd_main_512}. shows the BD-rate saving of five basic GFVC algorithms against VVC on both 256$\times$256 and 512$\times$512 resolution, it can be seen that DAC~\cite{ultralow} can achieve the highest BD-rate saving among all compared algorithms with more than 70\% BD-rate saving, followed by CFTE~\cite{CHEN2022DCC} and FV2V~\cite{wang2021Nvidia}, with more than 50\% BD-rate saving. TPS~\cite{9880299} and FOMM~\cite{FOMM} show less advantages due to their higher bit-rates.

For pixel-level metrics, the RD performance comparison for both 256$\times$256 and 512$\times$512 resolution are shown in Fig.~\ref{benchmark_main_result_pixel}. It can be observed that, for both Rate-PSNR and Rate-SSIM, GFVC algorithms fall far behind the VVC anchor and show obvious saturation. This is largely because GFVC algorithms are optimized for perceptual-level quality instead of pixel-level quality~\cite{chen2023generative}, and pixel-level alignment is not guaranteed under ultra-low bit-rate situation.

\begin{figure}[t]
\centering
\subfloat[Rate-DISTS]{\includegraphics[width=0.248\textwidth]{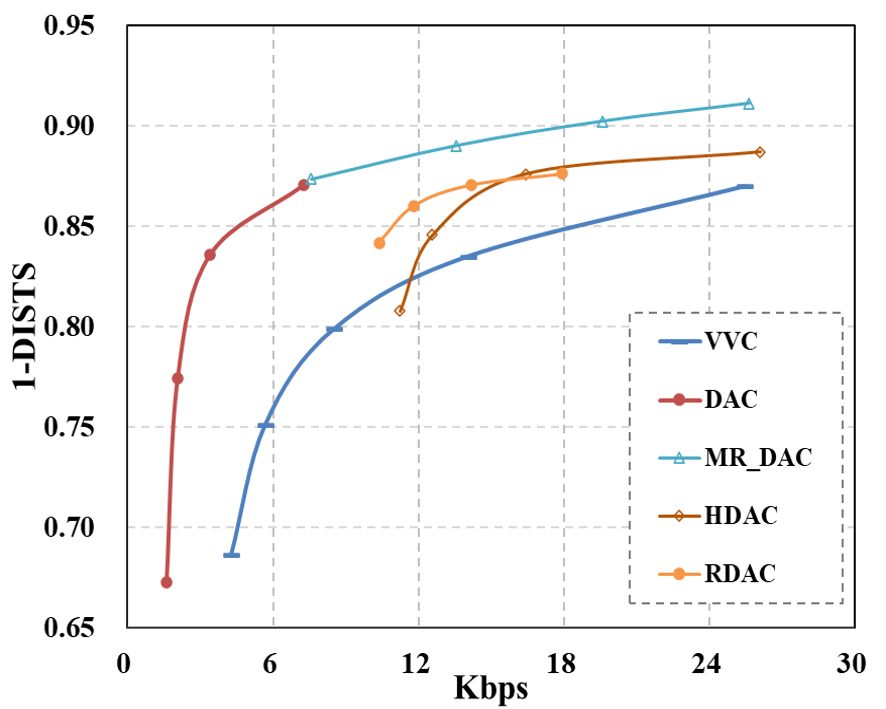}}
\subfloat[Rate-LPIPS]
{\includegraphics[width=0.248\textwidth]{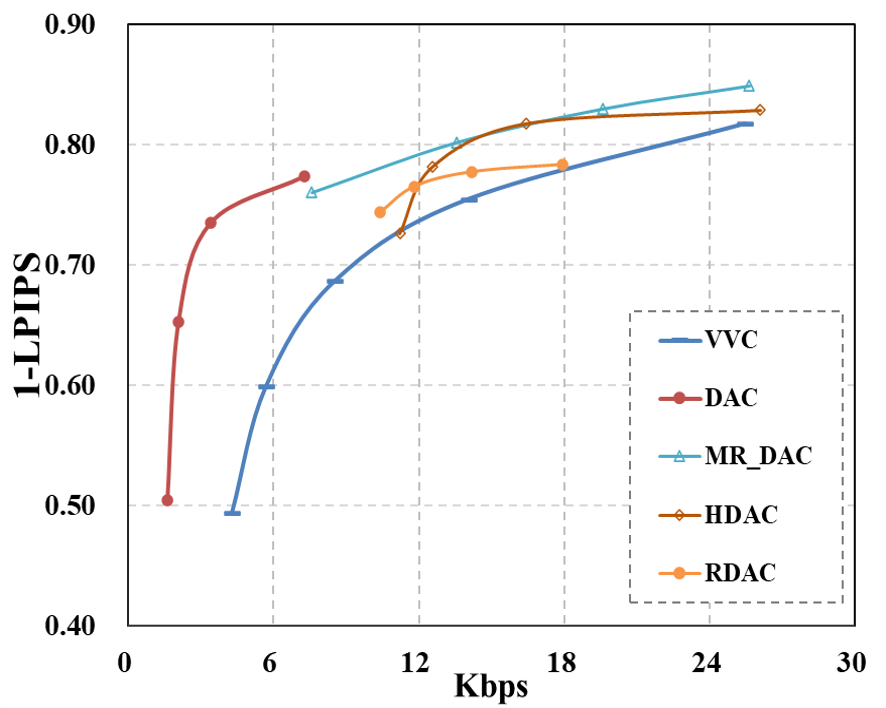}}
\\
\subfloat[Rate-PSNR]{\includegraphics[width=0.248\textwidth]{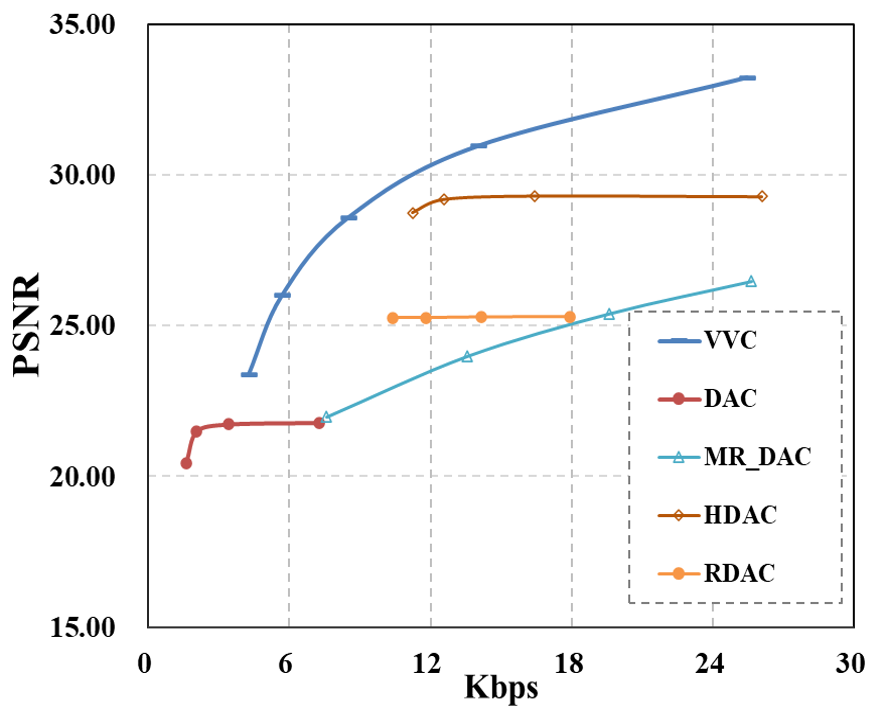}}
\subfloat[Rate-SSIM]
{\includegraphics[width=0.248\textwidth]{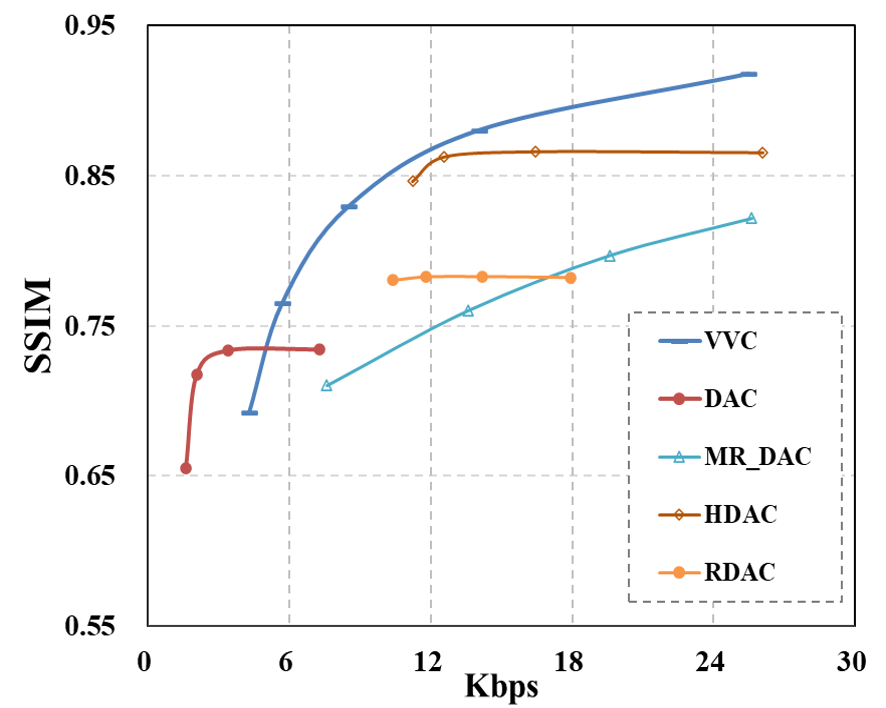}}

\caption{Rate-Distortion performance comparisons of VVC~\cite{bross2021overview}, DAC~\cite{ultralow} and its three further optimized models~(HDAC~\cite{konuko2022hybrid}, RDAC~\cite{konuko2023predictive}, MRDAC~\cite{MRDAC}) in terms of rate-DISTS and rate-LPIPS on 256$\times$256 resolution.} %
\label{benchmark_ext_result}
\end{figure}

\begin{table}[t]
\caption{RD performance comparisons on three further optimized GFVC algorithms in terms of average BD-rate savings over the VVC~\cite{bross2021overview} anchor on 512$\times$512 resolution}
    \label{bd_ext_256}
    \centering
\renewcommand\arraystretch{1.1}
\begin{tabular}{ccc}
\hline
Algorithm      & Rate-DISTS & Rate-LPIPS \\ \hline
HDAC~\cite{konuko2022hybrid}         & -11.14\%            & -17.39\%                      \\
RDAC~\cite{konuko2023predictive}           & -26.43\%            & -5.53\%                      \\
MRDAC~\cite{MRDAC}           & -28.13\%            & -29.14\%                      \\
 \hline
\end{tabular}
\vspace{-0.2cm}
\end{table}

\begin{figure*}[!t]
\centering
\vspace{-1.5em}
\subfloat[FOMM~\cite{FOMM}, CFTE~\cite{CHEN2022DCC}, FV2V~\cite{wang2021Nvidia}, DAC~\cite{ultralow}, TPS~\cite{8099774}, VVC~\cite{bross2021overview} at 5kps]
{\includegraphics[height=4.3cm]{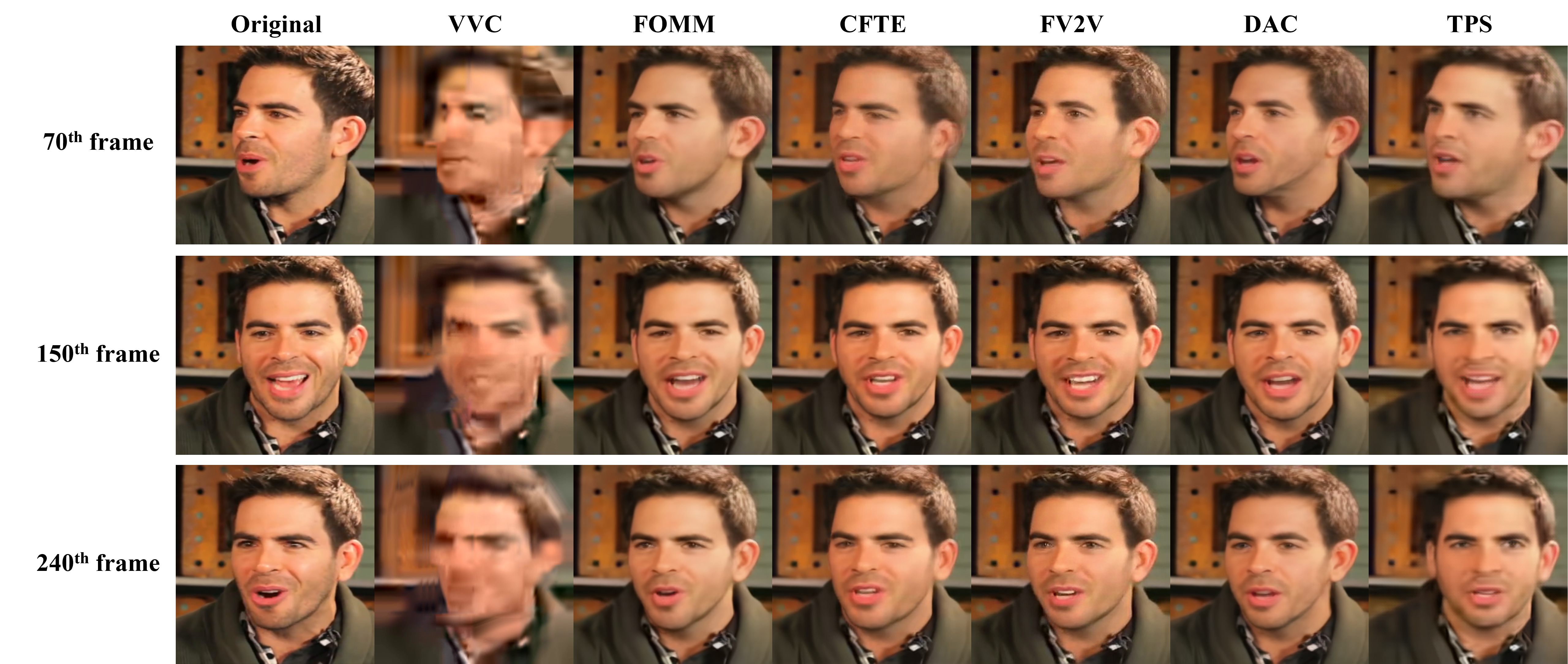}} 
\subfloat[HDAC~\cite{konuko2022hybrid}, RDAC~\cite{konuko2023predictive}, MRDAC~\cite{MRDAC}, VVC~\cite{bross2021overview} at 14kps]{\includegraphics[height=4.3cm]{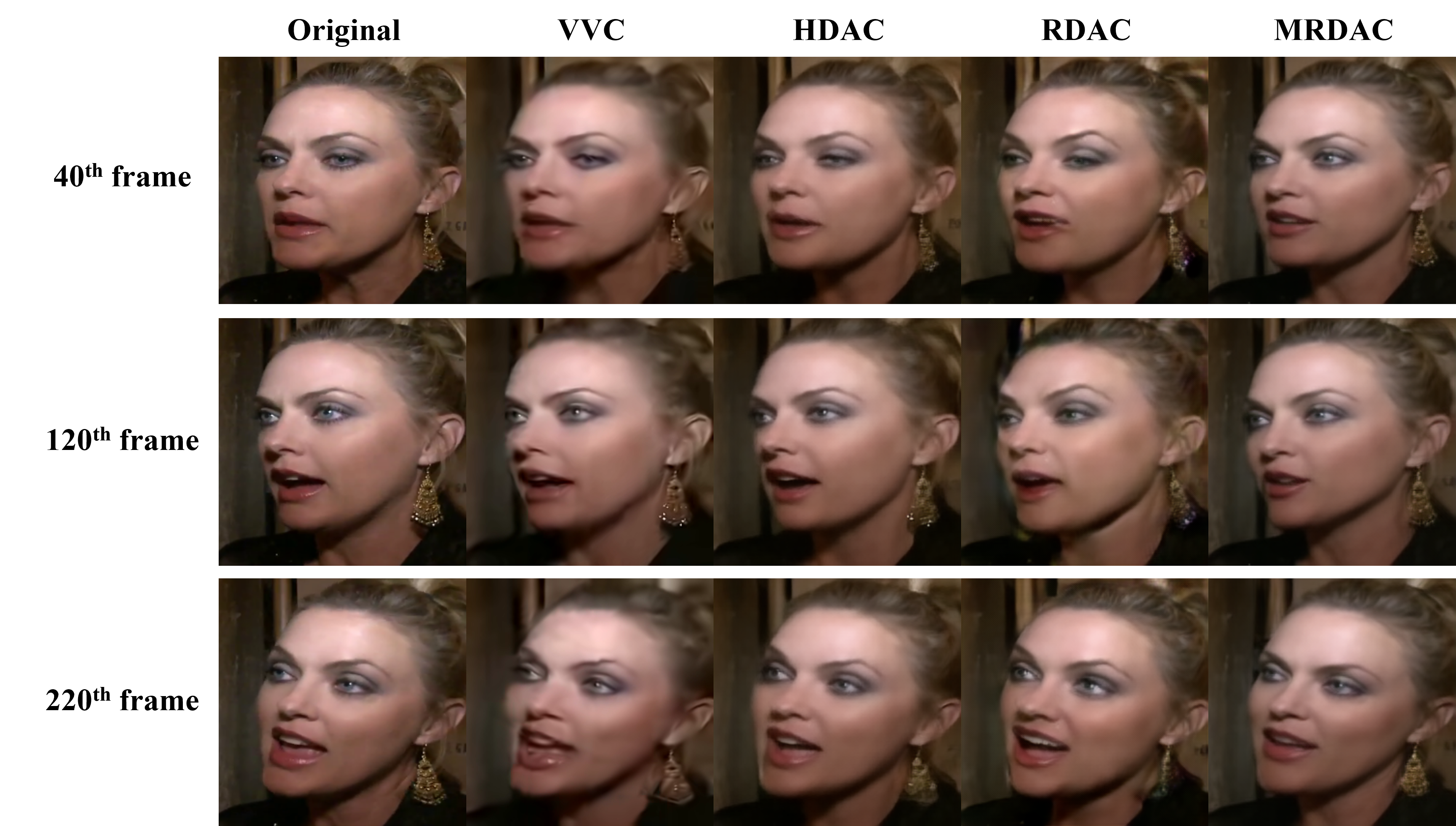}}
 \caption{Visual quality comparisons of GFVC algorithms and VVC~\cite{bross2021overview} at the similar bit rate.}
 \label{benchmark_subjective}
\end{figure*}

\textbf{b}) \textit{\textbf{Optimized GFVC Algorithms}}:\quad The RD performance of three further optimized GFVC algorithms in terms of both perceptual-level and pixel-level qualities are shown in Fig.~\ref{benchmark_ext_result}. It can be observed that all three algorithms can outperform VVC anchor in terms of Rate-DISTS and Rate-LPIPS. Specifically, MRDAC~\cite{MRDAC} can effectively extend the bit-rate coverage of DAC~\cite{ultralow} by including more key frames, and HDAC~\cite{konuko2022hybrid} can improve the reconstruction quality of VVC by fusing VVC reconstruction with GFV generation. For RDAC~\cite{konuko2023predictive} with additional residuals, it obtains less performance gain compared to MRDAC~\cite{MRDAC}, due to the difficulty of residual prediction and transmission. TABLE~\ref{benchmark_ext_result} shows the BD-rate savings over VVC anchor, it can be seen that MRDAC can achieve more than 30\% BD-rate saving in terms of Rate-DISTS and Rate-LPIPS, followed by RDAC achieving 26.43\% BD-rate saving on Rate-DISTS and HDAC achieving 17.39\% on Rate-LPIPS. 

For pixel-level performances,  HDAC~\cite{konuko2022hybrid} shows the closest quality to VVC anchor, indicating that using VVC~\cite{bross2021overview} as a base layer and building layered coding can improve the pixel-level alignment. MRDAC~\cite{MRDAC} provides the most sustained performance increasing from DAC~\cite{ultralow}. In particular, for both perceptual and pixel performance levels, MRDAC~\cite{MRDAC} shows the least trend of performance saturation and the most similar performance growth compared to the VVC anchor, illustrating the effectiveness and robustness of the multi-reference scheme with bidirectional prediction for GFVC optimization. 

\subsubsection{Subjective Performances} 

The subjective quality of compared GFVC algorithms are shown in Fig.~\ref{benchmark_subjective} , where basic GFVC algorithms are compared at ultra-low bit-rate, and optimized GFVC algorithms are compared at medium bit-rate.
It can be seen that, at ultra-low bit-rate, VVC reconstruction exhibits severe blocking effects and is almost unrecognizable, while GFVC algorithms can deliver more visually-pleasing reconstructions. Among all ultra-low bit-rate reconstructions, TPS~\cite{8099774} reconstructions are more blurry due to its inefficient feature representations, and CFTE~\cite{CHEN2022DCC} exhibits loss of details with larger head movements. For optimized GFVC algorithms, they all show clearer reconstructions than VVC while preserving more accurate details than basic GFVC algorithms. These results show the superior ability of GFVC algorithms to achieve vivid reconstructions under ultra-low bit-rate constraint as well as the potential to maintain their advantages in wider bit-rate ranges.


\section{Perceptual Quality Assessment on GFVC}
As mentioned in Section~\ref{sec:CTC}, perceptual objective metric are preferred for quality assessment of GFVC reconstructions, and DISTS, LPIPS have become de facto measurements in GFVC studies. However, there lacks a comprehensive comparison and evidence for selecting appropriate objective quality assessment methods for GFVC-compressed contents.
This section establishes a comprehensive GFVC-compressed face video database with human perception scores, and provides an in-depth analysis of the correlations between the mainstream objective quality measures and GFVC perception scores. 

\subsection{Subjective Quality Assessment}

\quad \textbf{1}) \textit{\textbf{GFVC-compressed Database}}:\quad We construct GFVC-compressed database using sequences from the benchmark evaluation in Section~\ref{benchmark_result}. Specifically, 33 testing sequences of 256$\times$256 resolution are compressed by eight different GFVC algorithms~(\textit{i.e.,} FOMM~\cite{FOMM}, CFTE~\cite{CHEN2022DCC}, FV2V~\cite{wang2021Nvidia}, DAC~\cite{ultralow}, TPS~\cite{8099774}, HDAC~\cite{konuko2022hybrid}, RDAC~\cite{konuko2023predictive} and MRDAC~\cite{MRDAC}), with four different qualities, yielding total 1056 sequences.

\textbf{2}) \textit{\textbf{Subjective Testing}}:\quad Following the common practice in the area of multimedia quality assessment~\cite{iqa_study,itu_bt_500}, we adopt the Double-Stimulus Continuous Quality Evaluation (DSCQE)~\cite{itu_bt_500} approach with the 5-category discrete Absolute Rating Scale (ARS)~\cite{ars} in our test. Totally 20 subjects participated in the testing, and all sequences are divided into four session with each lasting no more than an hour.

\textbf{3}) \textit{\textbf{Subjective Ratings Processing}}:\quad To guarantee cognitive impenetrability (i.e., consistent decision-making across individuals for varied video content), we perform agreement testing and subject rejection. Specifically, we exclude the sequences according to InterQuartile Range~(IQR)~\cite{li2023perceptual}, and we reject subjects if more than 5\% of their scores fall outside the range of 4.47 standard deviations from the mean scores~\cite{iqa_study}. During this process, 148 sequences are excluded and 1 subject is rejected. Then, reliable subjective ratings are subsequently transformed into the Mean Opinion Score (MOS) after they are first converted into Z-scores and then mapped to [0,100] following~\cite{iqa_study}.


\subsection{Performance of Objective Quality Assessment Methods}

\begin{table}[t]\scriptsize
\caption{Performance of eighteen objective quality assessment methods, where six best performed methods are underlined. KRCC and SRCC are displayed as absolute values.}
    \label{obj_performance}
    \centering
\renewcommand\arraystretch{1.35}
\resizebox{1.0\linewidth}{!}{
\begin{tabular}{cccccc}

\hline
Category & Algorithm      & PLCC~$\uparrow$ & SRCC~$\uparrow$ & KRCC~$\uparrow$  & RMSE~$\downarrow$\\ \hline
\multirow{5}{*}{\begin{tabular}[c]{@{}c@{}}Pixel \\ Level\end{tabular}}     &   PSNR~\cite{2009Mean}         & 0.52           & 0.50            & 0.34  &  11.67     \\
        &SSIM~\cite{2004Image}           & 0.53            & 0.51            & 0.35   &  11.61     \\
        &VMAF~\cite{vmaf}           & 0.72            & 0.64           & 0.46   &    9.49   \\
        &FSIM~\cite{fsim}            & 0.65           & 0.62            & 0.45    &  10.37   \\
        &VSI~\cite{vsi}            & 0.59          & 0.56            & 0.39   & 11.01     \\
        \hline
\multirow{4}{*}{\begin{tabular}[c]{@{}c@{}}Perceptual \\ Level\end{tabular}}   &      DISTS~\cite{dists}            & \textbf{\uline{0.86}}           & \textbf{\uline{0.80}}            & \textbf{\uline{0.62}}    &   \textbf{\uline{6.92}}  \\
        &LPIPS~\cite{lpips}            & \textbf{\uline{0.85}}           & \textbf{\uline{0.80}}           & \textbf{\uline{0.61}}  &  \textbf{\uline{7.12}}    \\
        &MSVGG~\cite{msvgg}            & 0.80            & 0.73           & 0.54    & 8.17    \\
        &TOPIQ~\cite{topiq}            & \textbf{\uline{0.86}}           & \textbf{\uline{0.81}}           & \textbf{\uline{0.62}}      & \textbf{\uline{6.91}}   \\
        \hline
\multirow{4}{*}{\begin{tabular}[c]{@{}c@{}}Face \\ Oriented\end{tabular}}  &      FAVOR~\cite{li2023perceptual}            & \textbf{\uline{0.81}}             & \textbf{\uline{0.74}}             & \textbf{\uline{0.55}}   &   \textbf{\uline{8.04}}     \\
        &IFQA~\cite{ifqa}            & 0.65           & 0.57            & 0.41     & 10.32   \\
        &AKD~\cite{FOMM}            & 0.44            & 0.41            & 0.29    &  12.24    \\
       & AED~\cite{FOMM}            &0.62           & 0.59            & 0.41    & 10.75     \\
        \hline
        
\multirow{2}{*}{\begin{tabular}[c]{@{}c@{}}Temporal \\ Consistency\end{tabular}}    &     FVD~\cite{Unterthiner2019FVDAN}            & \textbf{\uline{0.85}}           & \textbf{\uline{0.82}}          & \textbf{\uline{0.63}}     &  \textbf{\uline{7.20}}   \\
        &MotionFlow~\cite{chen2023csvt}            & 0.20           & 0.22            & 0.14 & 13.38       \\
        \hline
\multirow{3}{*}{\begin{tabular}[c]{@{}c@{}}No \\ Reference\end{tabular}}    &     MANIQA~\cite{yang2022maniqa}            & \textbf{\uline{0.84}}            & \textbf{\uline{0.78}}           & \textbf{\uline{0.59}}    &  \textbf{\uline{7.34}}   \\
        &BRISQUE~\cite{BRISQUE}            & 0.57           & 0.52          & 0.37    & 11.26     \\
        &STRRED~\cite{STRRED}            & 0.60           & 0.57          & 0.40     &  10.92  \\
        \hline
\end{tabular}
}
\end{table}

\begin{figure}[t]
\centering
\subfloat[DISTS]{\includegraphics[width=0.16\textwidth]{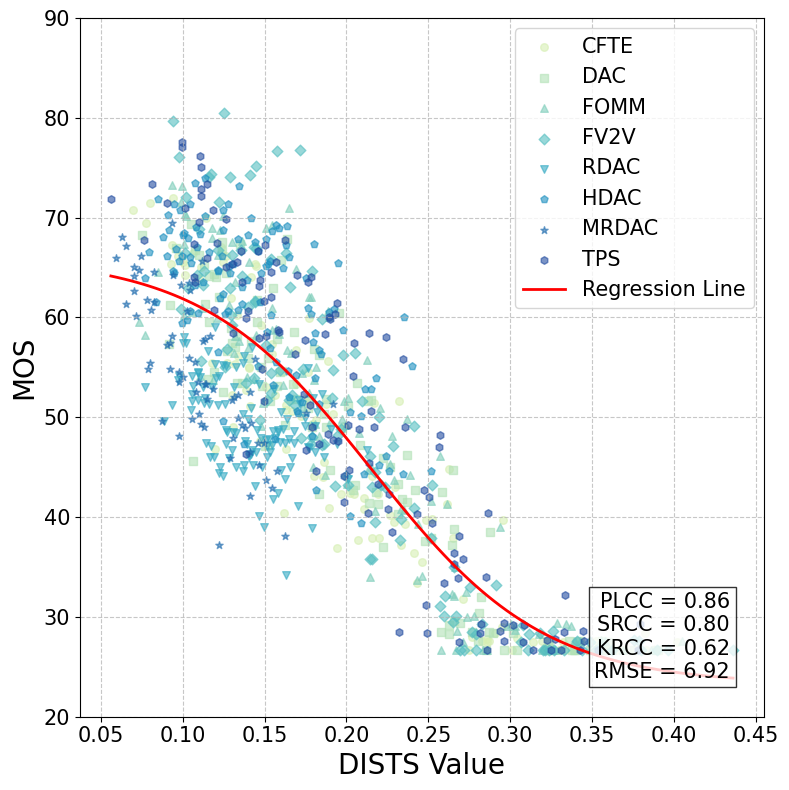}}
\subfloat[TOPIQ]
{\includegraphics[width=0.16\textwidth]{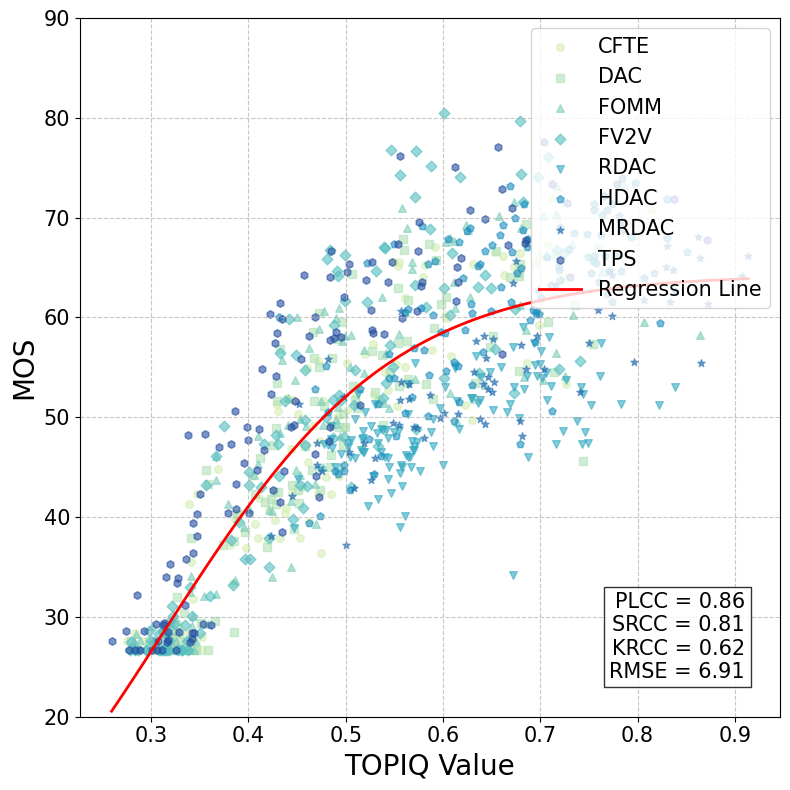}}
\subfloat[LPIPS]{\includegraphics[width=0.16\textwidth]{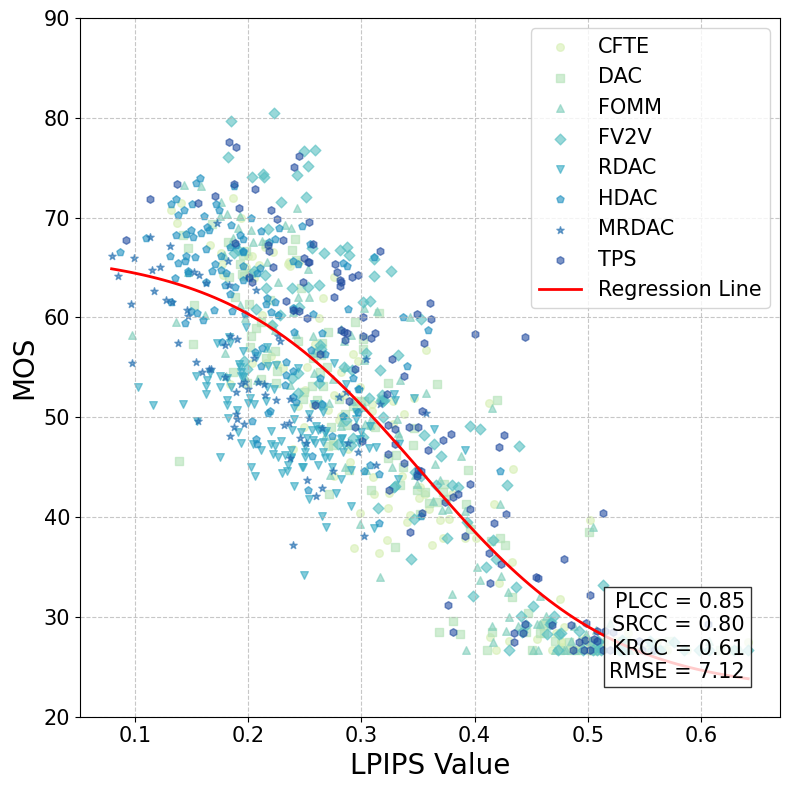}}
\\
\subfloat[FVD]
{\includegraphics[width=0.16\textwidth]{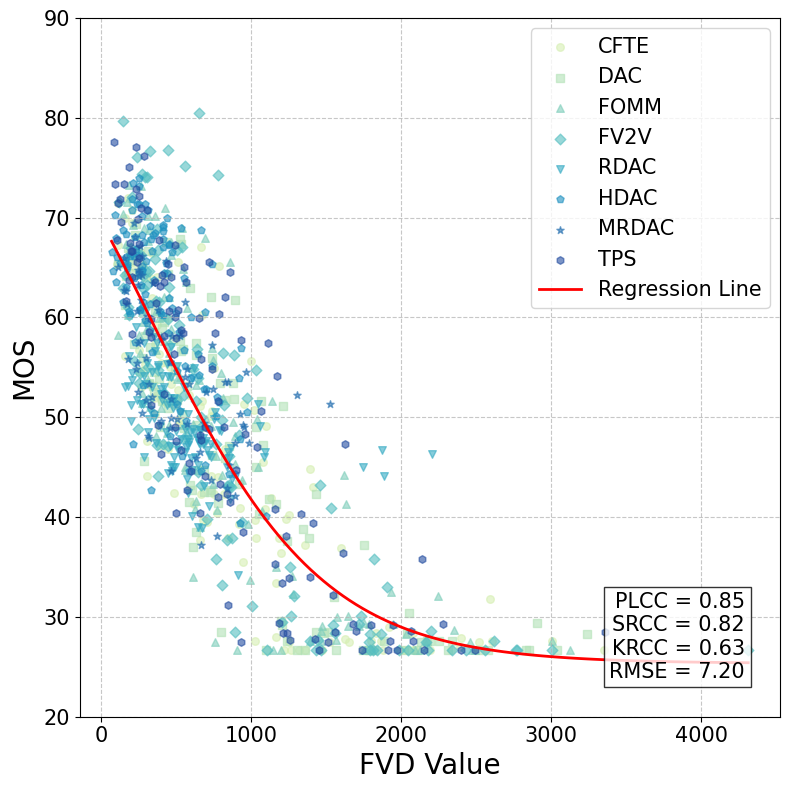}}
\subfloat[MANIQA]{\includegraphics[width=0.16\textwidth]{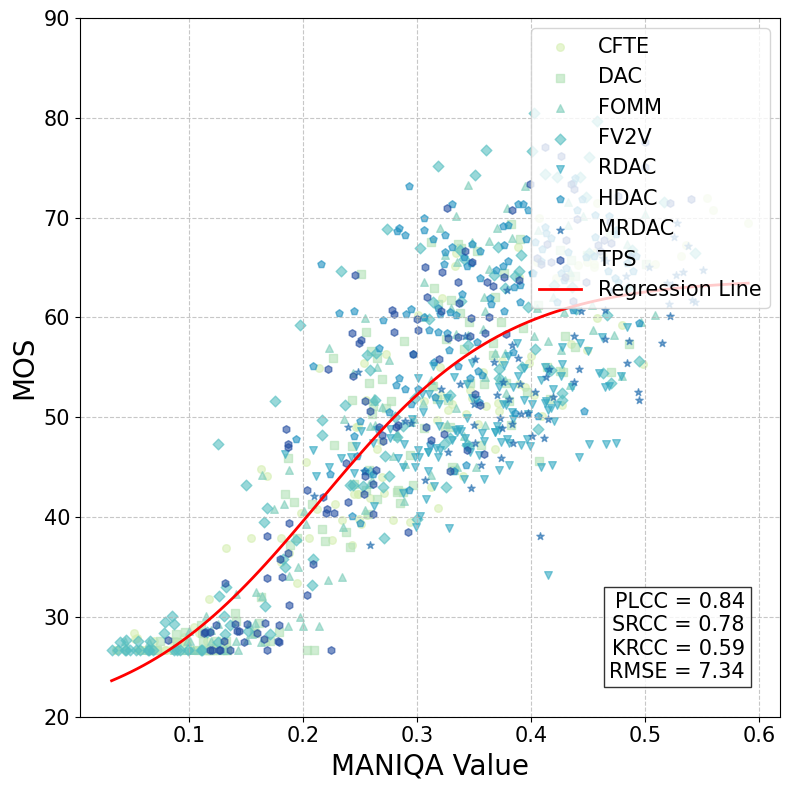}}
\subfloat[FAVOR]
{\includegraphics[width=0.16\textwidth]{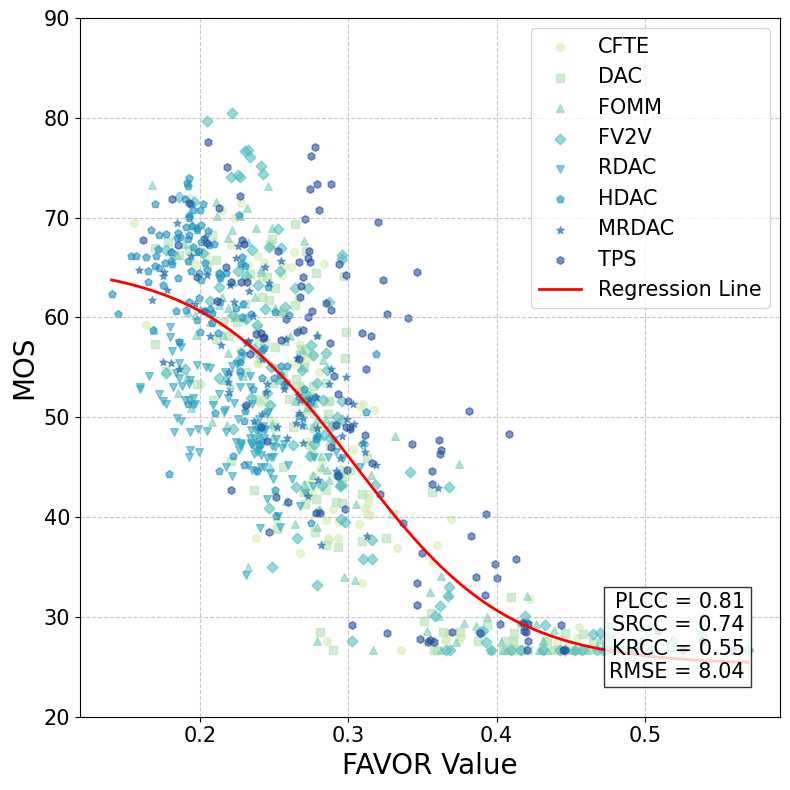}}
\caption{Comparison of human mean opinion scores (MOSs) against DISTS~\cite{dists}, TOPIQ~\cite{topiq}, LPIPS~\cite{lpips}, FVD~\cite{Unterthiner2019FVDAN}, MANIQA~\cite{yang2022maniqa} and FAVOR~\cite{li2023perceptual} on the GFVC-compressed database.} %
\label{iqa_fitting}
\end{figure}

To evaluate the performances of difference objective quality methods on GFVC-compressed content, we select 18 metrics and evaluate their correlation with subjective quality. Objective image/video quality assessment methods with different design philosophy are considered. The details are listed as follows,
\begin{itemize}
\item \textbf{Pixel-level Metrics} directly apply mathematical calculation on raw pixels: 
PSNR~\cite{2009Mean} calculates the pixel-wise mean square error between two images. SSIM~\cite{2004Image} evaluates the structural similarity within a local window. Video Multimethod Assessment Fusion~(VMAF)~\cite{vmaf} extracts visual quality fidelity and detail loss in pixel neighborhood and then measure temporal deviation across frames. Feature Similarity Index Measure~(FSIM)~\cite{fsim} utilizes phase congruency and gradient magnitude to obtain low-level features. Visual Saliency-Induced Index~(VSI)~\cite{vsi} further introduces visual saliency to indicate the local quality.

\item \textbf{Perceptual-level Metrics} utilize deep features from pre-trained neural network to provide more semantic perception: MSVGG~\cite{msvgg} is first proposed for training losses by calculating L1 norm of pre-trained VGG network features differences, then the effectiveness of deep features as perceptual metrics is comprehensively discussed in~\cite{lpips} and LPIPS is proposed as weighted-sum L2 norm of VGG feature differences. Similarly, DISTS~\cite{dists} measures and fuses the structural similarity of VGG features. Instead of bottom-up fusion of multi-scale VGG feature differences, TOPIQ~\cite{topiq} leverages semantic propagation in a top-down manner to distinguish the semantically important local distortions.

\item \textbf{Face-oriented Metrics} are specially designed for human-face-centric contents: Interpretable Face Quality Assessment~(IFQA)~\cite{ifqa} builds an adversarial framework to generate simulated distortion and discriminate face quality. FAce
VideO IntegeRity~(FAVOR)~\cite{li2023perceptual} considers distinct content characteristics and temporal priors of face videos, and optimizes its framework on a compressed face database. Besides, Average Keypoint Distance~(AKD) and Average Euclidean Distance~(AED) that are widely adopted for face animation evaluation~\cite{FOMM} are also included.

\item \textbf{Temporal Consistency Metrics} focus on the inter-frame distortion of videos: Fréchet Video Distance~(FVD)~\cite{Unterthiner2019FVDAN} captures the temporal dynamics and compares the feature distribution, and MotionFlow is proposed as a training loss in~\cite{chen2023csvt} by comparing the optical flow of adjacent frames extracted from SpyNet~\cite{spynet}. 

\item \textbf{Non-reference Metrics}: In addition to the above-mentioned full-reference metrics, three no-reference metrics are also evaluated, including Multi-dimension Attention Network for no-reference Image Quality Assessment~(MANIQA)~\cite{yang2022maniqa}, Blind/Referenceless Image Spatial Quality Evaluator~(BRISQUE)~\cite{BRISQUE} with scene statistics of locally normalized luminance coefficients, and Spatio-temporal Reduced Reference Entropic Differences~(STRRED)~\cite{STRRED}.
\end{itemize}

To evaluate the performance of selected objective quality assessment metrics, Pearson Linear Correlation Coefficient~(PLCC), Spearman Rank Correlation Coefficient~(SRCC), Kendall Rank Correlation Coefficient~(KRCC), and Root-Mean-Square Error~(RMSE) are used to evaluate the correlation between objective measurements and subjective MOSs. Before computing the PLCC and RMSE, objective quality scores are remapped by fitting the non-linear logistic regression function,
\begin{equation}
    f(x) = \frac{(\eta_{1} - \eta_{2})}{1+e^{\frac{-(x-\eta_{3})}{|\eta_{4}|}}} + \eta_{2},
\end{equation}
where $\{ \eta_{i} \}_{i=1}^{4}$ are fitting parameters. 

TABLE~\ref{obj_performance} summarizes the performance of objective quality assessments methods on the GFVC-compressed database, where KRCC and SRCC are displayed as absolute values. It can be seen that DISTS~\cite{dists}, TOPIQ~\cite{topiq}, FVD~\cite{Unterthiner2019FVDAN} and LPIPS~\cite{lpips} are able to obtain the best performances with PLCC, SRCC over 0.8, kRCC over 0.6 and RMSE below 8. Besides, MANIQA, FAVOR and MSVGG can also achieve PLCC, SRCC over 0.7 and KRCC over 0.5. The other metrics perform less satisfactory with MotionFlow~\cite{chen2023csvt} showing the worst performances with correlations around 0.2.

Among all categories of objective metrics, perceptual-level metrics are able to achieve the most promising results, indicating that the generative video coding that relies more on semantic reconstruction to fulfill human perception under ultra-low bit-rate constraint is built on solid foundation. On the contrary, traditional pixel-level metrics appear to be less correlated with human perception and less suitable for quality assessment for GFVC-compressed content. Other advantageous metrics include face-oriented FAVOR~\cite{li2023perceptual}, temporal measurement FVD~\cite{Unterthiner2019FVDAN} and non-reference MANIQA~\cite{yang2022maniqa}. Despite their advantages among all comparison metrics, correlations around 0.85 are still not strong enough to match the human opinions. We display the scatter plots of the objective quality scores and MOSs of six advantageous metrics in Fig.~\ref{iqa_fitting}, and we can still see obvious disagreement between objective and subjective scores.
Specifically, on the lower left corner of Fig.~\ref{iqa_fitting}.(a),(c),(d),(f) and upper right corner of Fig.~\ref{iqa_fitting}.(b)(e), the video samples are good for objective metrics but bad for human perception.
For example, RDAC~\cite{konuko2023predictive} samples~(inverted triangle markers) are quite common in those contradictory areas, which is possibly caused by temporal flickering from its GOP-based residual refreshing. Using predictive residuals helps improving objective quality but the temporal discontinuity could be annoying for subjective perception.
It indicates that a more suitable objective quality assessment metric is required for generative reconstructed video content.

\section{GFVC Standardization Efforts and Low-complexity System Implementation}
This section reviews the GFVC's standardization efforts in the past years and provides a low-complexity GFVC system implementation with model lightweight techniques, demonstrating great possibilities in future deployments and applications.

\subsection{Standardization Efforts with Supplementary Enhancement Information}

\subsubsection{Overview}

\begin{table*}[!t]
\renewcommand\arraystretch{1.5}
\caption{Summary of GFVC standardization efforts made by JVET}  
\label{table2}
\centering
\resizebox{1\textwidth}{!}{
\begin{tabular}{lllm{7.8cm}} 
\hline
\textbf{Document Number}                                                                                                                & \textbf{Proposal Time}              & \textbf{Related Topic}           & \textbf{Main Contribution}                                                                                                                                                                                          \\ \hline
JVET-AC0088~\cite{JVET-AC0088}                                                                                                                    & 2023 Jan                   & SEI Messages            & Proposes the first Generative Face Video (GFV) SEI message to allow VVC-coded pictures to be used as base pictures and add a small amount of bit overhead to represent facial semantics.                   \\
JVET-AD0051~\cite{JVET-AD0051}                                                                                                                     & 2023 Apr                   & SEI Messages            & Generalizes the GFV SEI message to include syntax elements that correspond to various facial representations.                                                                                              \\
JVET-AE0080~\cite{JVET-AE0080}/JVET-AE0088~\cite{JVET-AE0088}/JVET-AE0280~\cite{JVET-AE0280}                                                                                            & 2023 Jul                   & SEI Messages            & Support more common syntax design of GFV SEI message and define the interface between the decoder (which decodes the reference pictures and parses the SEI message) and the generative neural network. \\
JVET-AE0083~\cite{JVET-AE0083}/JVET-AH0239~\cite{JVET-AH0239}                                                                                                        & 2023 Jul/2024 Apr        & Software Tools & Provides clear information on available open-source generative models as well as an example software implementation that could link the VTM software with a generative network.                          \\
JVET-AF0048~\cite{JVET-AF0048}                                                                                                                    & 2023 Oct                   & Software Tools  & Uses a flow translator or a parameter translator to improve the interoperability between encoder and decoder that have mismatched facial feature type.                                                     \\
JVET-AF0146~\cite{JVET-AF0146}                                                                                                                    & 2023 Oct                   & SEI Messages            & Soughts to introduce refinements and grouping of 3D facial landmarks syntax for the GFV SEI message.                                                                                                       \\
JVET-AF0234~\cite{JVET-AF0234}                                                                                                                    & 2023 Oct                   & SEI Messages            & Updates the common text of GFV SEI message proposed in JVET-AE0280 by introducing the parameter translator proposed in JVET-AF0048.                                                                        \\
JVET-AG0042~\cite{JVET-AG0042}                                                                                                                     & 2024 Jan                   & Software Tools   & Proposes common software tools and testing conditions for Generative Face Video Coding.                                                                                                               \\
JVET-AG0048~\cite{JVET-AG0048}/JVET-AL0147~\cite{JVET-AL0147}                                                                                                       & 2024 Jan/2025 Apr          & Software Tools  & Develop parameter translator networks to allow ``mismatched'' encoder and decoder to interoperate.                                                                                                           \\
JVET-AG0087~\cite{JVET-AG0087}/JVET-AG0088~\cite{JVET-AG0088}/JVET-AG0203~\cite{JVET-AG0203}                                                                                            & 2024 Jan                   & SEI Messages            & Fix GFV SEI bugs and defines the generator neural network for GFV.                                                                                                                                         \\
JVET-AG0139~\cite{JVET-AG0139}                                                                                                                     & 2024 Jan                   & Software Tools       & Leverages depthwise separable convolution to lightweight the GFVC model.                                                                                                                                   \\
JVET-AG0187~\cite{JVET-AG0187}/JVET-AG2035~\cite{JVET-AG2035}/JVET-AJ2035~\cite{JVET-AJ2035}                                                                                            & 2024 Jan/2024 Nov         & Test Conditions         & Constructs test conditions and evaluation procedures for generative face video coding.                                                                                                                     \\
JVET-AH0053~\cite{JVET-AH0053}/JVET-AH0054~\cite{JVET-AH0054}/JVET-AH0148~\cite{JVET-AH0148}                                                                                            & 2024 Apr                   & SEI Messages            & Fixes GFV SEI bugs and improve the syntax text quality.                                                                                                                                                    \\
JVET-AH0109~\cite{JVET-AH0109}                                                                                                                    & 2024 Apr                   & Software Tools       & Removes the analysis model on the decoder side and lightweights the GFVC models.                                                                                                                           \\
JVET-AH0110~\cite{JVET-AH0110}                                                                                                                    & 2024 Apr                   & Software Tools      & Propose a scalable   representation and layered reconstruction architecture to extend the bitrate and quality range that GFVC methods in AHG16 software can cover.                                         \\
JVET-AH0113~\cite{JVET-AH0113}                                                                                                                    & 2024 Apr                   & Software Tools      & Develops a multi-resolution support scheme for GFVC models                                                                                                                                                 \\
JVET-AH0114~\cite{JVET-AH0114}                                                                                                                    & 2024 Apr                   & Software Tools   & Incorporates two new GFVC algorithms into the AHG16 software and improves the AHG16 GFVC software code.                                                                                                    \\
JVET-AH0118~\cite{JVET-AH0118}                                                                                                                    & 2024 Apr                   & SEI Messages            & Describes a showcase for picture fusion for the GFV SEI message.                                                                                                                                           \\
JVET-AH0127~\cite{JVET-AH0127}                                                                                                                    & 2024 Apr                   & SEI Messages            & Proposes a new enhancement SEI message for generative face video.                                                                                                                                          \\
JVET-AH0138~\cite{JVET-AH0138}/JVET-AI0137~\cite{JVET-AI0137}/JVET-AJ0135~\cite{JVET-AJ0135}                                                                                            & 2024 Apr/2024 Jul/2024 Nov & SEI Messages            & Proposes to add 512x512 test sequences to the GFVC test conditions, and provides performance results using SEI message to code such 512x512 content following GFVC test   configurations.                  \\
JVET-AI0047~\cite{JVET-AI0047}/JVET-AJ0209~\cite{JVET-AJ0209}                                                                                                        & 2024 Jul/2024 Nov          & Test Conditions         & Develops a new set of higher resolution (512x512) test content for GFVC experimentation, and provides experimental results of 512x512 content following GFVC test   configurations.                        \\
JVET-AI0048~\cite{JVET-AI0048}                                                                                                                    & 2024 Jul                   & Software Tools      & Improves GFVC performance, esp. for head-and-shoulder content, by retraining the models with more diverse training data.                                                                                   \\
\begin{tabular}[c]{@{}l@{}}JVET-AI0156~\cite{JVET-AI0156}/JVET-AI0184~\cite{JVET-AI0184}/JVET-AI0189~\cite{JVET-AI0189}/\\ JVET-AI0190~\cite{JVET-AI0190}/JVET-AI0191~\cite{JVET-AI0191}/JVET-AI0193~\cite{JVET-AI0193}\end{tabular}             & 2024 Jul                   & SEI Messages            & Propose various bug fixes and refinements to GFV SEI and raise some topics for discussion.                                                                                                                  \\
JVET-AI0186~\cite{JVET-AI0186}/JVET-AI0192~\cite{JVET-AI0192}                                                                                                        & 2024 Jul                   & SEI Messages            & Add timing information to GFV SEI.                                                                                                                                                                         \\
JVET-AI0194~\cite{JVET-AI0194}                                                                                                                     & 2024 Jul                   & SEI Messages            & Adds chroma key information to GFV SEI.                                                                                                                                                                    \\
JVET-AI0195~\cite{JVET-AI0195}                                                                                                                    & 2024 Jul                   & SEI Messages            & Modifies syntax coding in order to reduce GFV SEI overhead.                                                                                                                                                \\
\begin{tabular}[c]{@{}l@{}}JVET-AJ0051~\cite{JVET-AJ0051}/JVET-AJ0069~\cite{JVET-AJ0069}/JVET-AJ0108~\cite{JVET-AJ0108}/\\ JVET-AJ0111~\cite{JVET-AJ0111}/JVET-AJ0132~\cite{JVET-AJ0132}\end{tabular}                         & 2024 Nov                   & SEI Messages            & Contribute to various high level syntax aspects of the GFV SEI message, including translator nerual network, instance count signalling, output order, etc.                                                 \\
JVET-AJ0052~\cite{JVET-AJ0052}                                                                                                                    & 2024 Nov                   & Software Tools   & Proposes multi-resolution FOMM, FV2V, CFTE, DAC models that can be used to code both 256x256 and 512x512 content.                                                                                          \\
JVET-AJ0207~\cite{JVET-AJ0207}                                                                                                                    & 2024 Nov                   & SEI Messages            & Illustrates the SEI overhead reduction effort, and proposes to move GFV and GFVE SEI messages from TuC to VSEI v4.                                                                                         \\
JVET-AK0068~\cite{JVET-AK0068}/JVET-AL0101~\cite{JVET-AL0101}                                                                                                        & 2025 Jan/2025 Apr          & Software Tools   & Propose a GFVC extension of the   VVC standard, which employs a generic video coding standard VVC for face video compression and does not need any additional SEI message in   transmission.               \\
JVET-AK0069~\cite{JVET-AK0069}/JVET-AL0102~\cite{JVET-AL0102}                                                                                                        & 2025 Jan/2025 Apr          & Software Tools   & Develop a QP-adaptive GFVC framework that incorporates varying feature sizes in SEI message according to the given QP value.                                                                               \\
\begin{tabular}[c]{@{}l@{}}JVET-AK0080~\cite{JVET-AK0080}/JVET-AK0127~\cite{JVET-AK0127}/JVET-AK0128~\cite{JVET-AK0128}/JVET-AK0154~\cite{JVET-AK0154}/\\ JVET-AK0164~\cite{JVET-AK0164}/JVET-AK0238~\cite{JVET-AK0238}/JVET-AK0239~\cite{JVET-AK0239}\end{tabular} & 2025 Jan                   & SEI Messages            & Semantics and syntax fixes for GFV SEI message and GFVE SEI message.                                                                                                                                       \\
JVET-AK0124~\cite{JVET-AK0124}                                                                                                                    & 2025 Jan                   & SEI Messages            & Proposes to add timing information and order of pictures in GFV SEI message.                                                                                                                               \\
JVET-AL0156~\cite{JVET-AL0156}                                                                                                                    & 2025 Apr                   & Software Tools      & Proposes to add a colour calibration process on the generated face pictures as an optional post-processing to solve the occasional colour shift issue.                                                     \\
JVET-AL0148~\cite{JVET-AL0148}                                                                                                                    & 2025 Apr                   & SEI Messages            & Proposes to include GFV and GFVE SEI messages for HEVC and AVC.                                                                                                                                            \\
JVET-AL0155~\cite{JVET-AL0155}                                                                                                                    & 2025 Apr                   & SEI Messages            & Further fixes and cleanup on GFV and GFVE SEI messages.                                                                                                                                                    \\ \hline
\end{tabular}
}
\end{table*}


Over the past 2.5 years, the standardization efforts of GFVC have made significant strides towards establishing a standardized framework for coding face video content using generative networks. As shown in TABLE \ref{table2}, there are over 60 technical proposals regarding GFVC techniques submitted to JVET. These efforts have focused on incorporating diverse GFVC characteristics via standardized high-level syntax, adding the corresponding syntax into VVC bitstreams, developing the reference software to link the VTM software with GFVC models, and investigating the interoperability and robustness of GFVC models. 

\begin{figure}[tb]
\centering
\centerline{\includegraphics[width=0.5\textwidth]{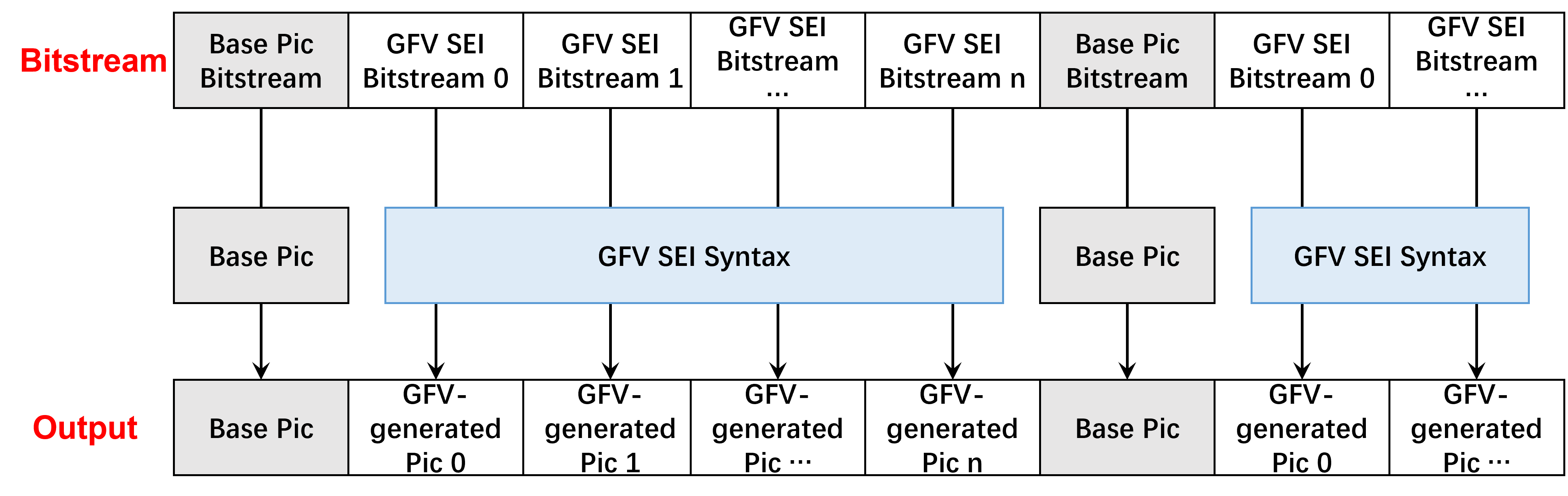}}  
\caption{Illustrations of SEI-based GFVC workflow at the decoder side. When receiving the hybrid coded bitstreams, the base picture will be reconstructed as texture reference and the facial parameters encapsulated into GFV SEI messages will be parsed to generate the animated face pictures. It should be noted that the base picture can be coded by all mainstream video coding standards like VVC, HEVC, and AVC.}
\label{fig_section5_workflow} 
\end{figure}

In particular, JVET has discussed adding the corresponding Supplementary Enhancement Information (SEI) into VVC bitstreams in order to support GFVC. As the name implies, SEI messages serve as extra data within a coded video bitstream, enriching the utility of transmitted video for diverse applications. The distinctive approach of GFVC, utilizing a hybrid codec and extracting facial parameters, presents significant potential for standardization as SEI messages within a video coding specification. As such, diverse GFVC features can be seamlessly inserted with the coded bitstreams from a traditional hybrid video codec, all without necessitating normative alterations to the encoder or decoder of the hybrid video codec. The detailed workflow of the SEI-based GFVC approach is shown in Fig. \ref{fig_section5_workflow}.

With respect to advances in GFVC standardization, JVET has established an Ad hoc Group (AhG) dedicated to in-depth investigations of GFVC, encompassing coordinated experimentation, interoperability study, establishing GFVC test conditions, and developing software tools. Furthermore, the Generative Face Video (GFV) and Generative Face Video Enhancement (GFVE) SEI messages have been incorporated into the draft amendment~\cite{JVET-AJ2006} of the Versatile Supplemental Enhancement Information (VSEI) standard, which will be standardized as a new version for ITU-T H.274 $|$ ISO/IEC 23002-7. Additionally, the GFV and GFVE SEI messages have recently been incorporated into the next versions~\cite{JVET-AL0148} of AVC~\cite{wiegand2003overview} and HEVC~\cite{sullivan2012overview} standards. In other words, the next versions of major video coding standards like VVC, HEVC, and AVC, will support the integration of GFV/GFVE SEI messages into their respective bitstreams. This achievement highlights the successful collaborative efforts of multiple companies and research institutions in standardizing generative video coding techniques within the JVET.

\subsubsection{Results Analysis}
\begin{figure}[t]
\centering
\subfloat[256$\times$256: Rate-DISTS]{\includegraphics[width=0.25\textwidth]{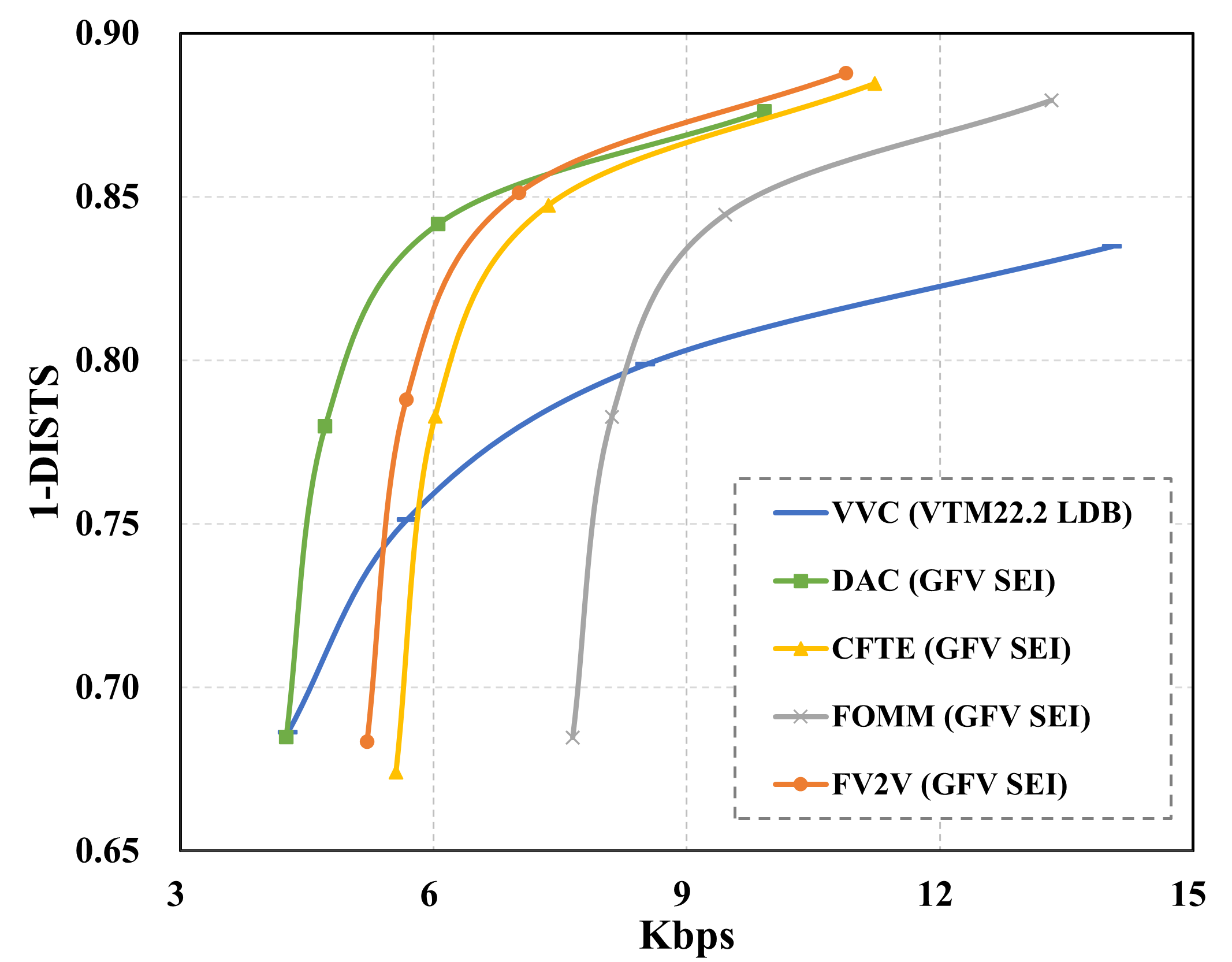}}
\subfloat[256$\times$256: Rate-LPIPS]{\includegraphics[width=0.25\textwidth]{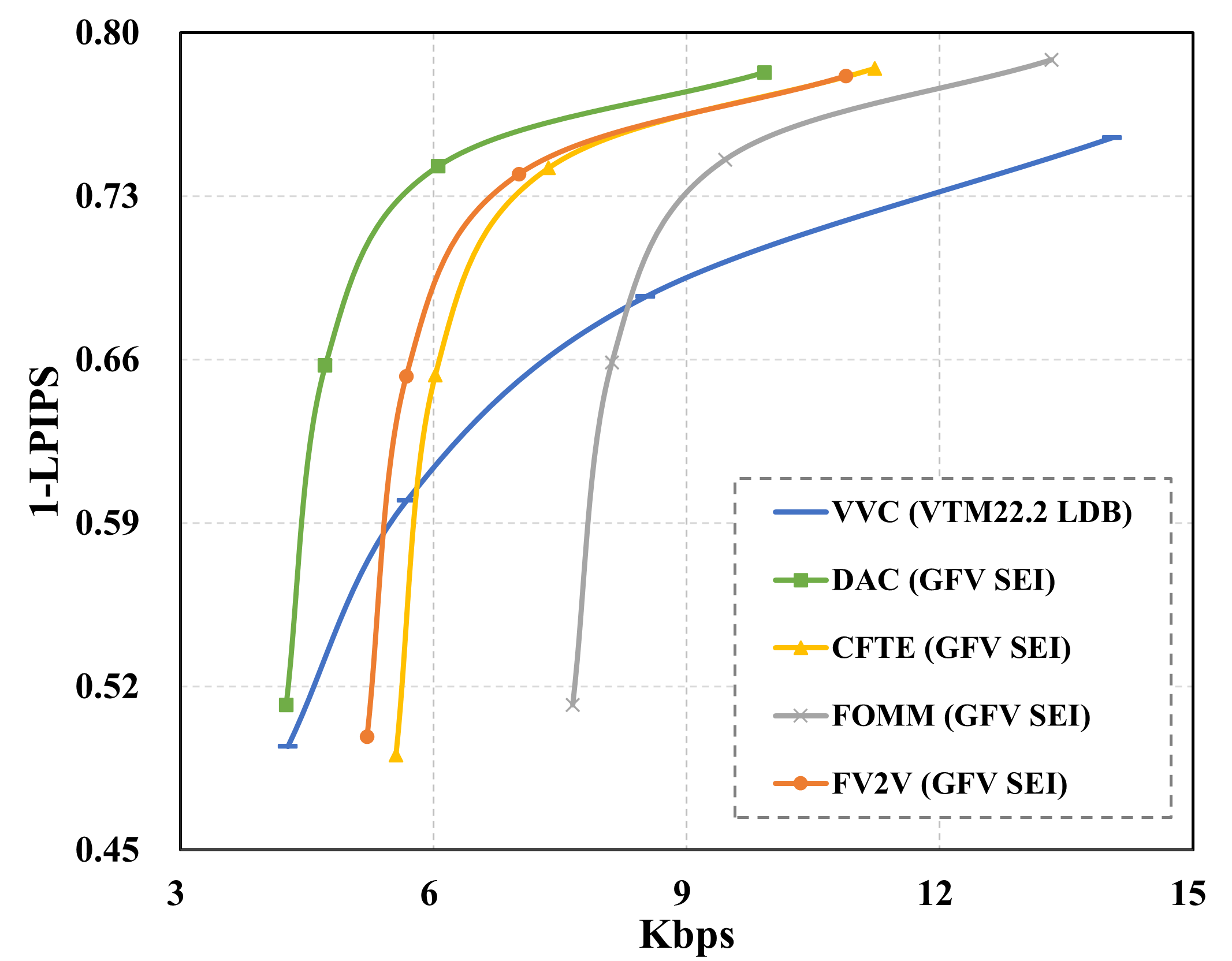}}\\
\subfloat[512$\times$512: Rate-DISTS]{\includegraphics[width=0.25\textwidth]{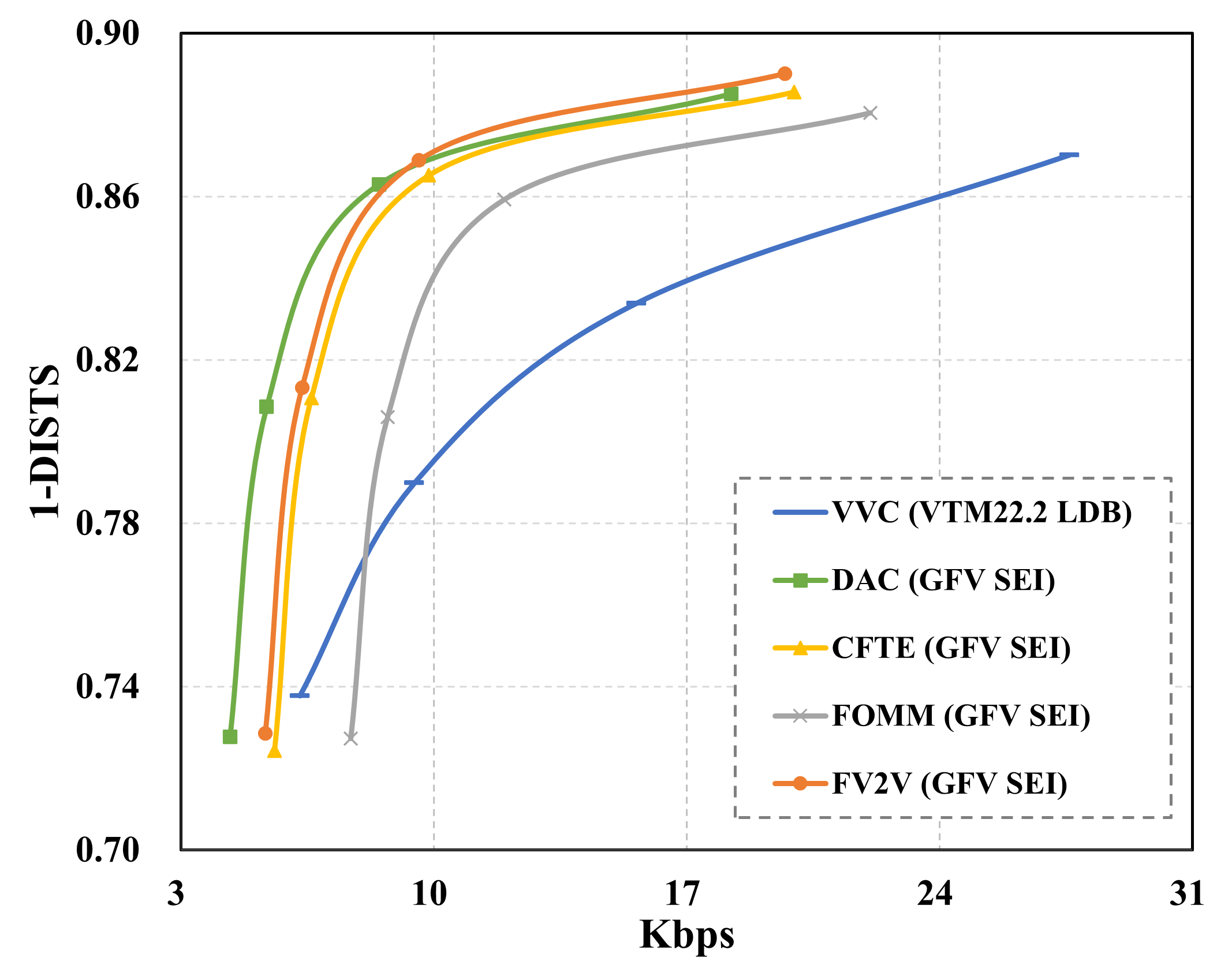}}
\subfloat[512$\times$512: Rate-LPIPS]{\includegraphics[width=0.25\textwidth]{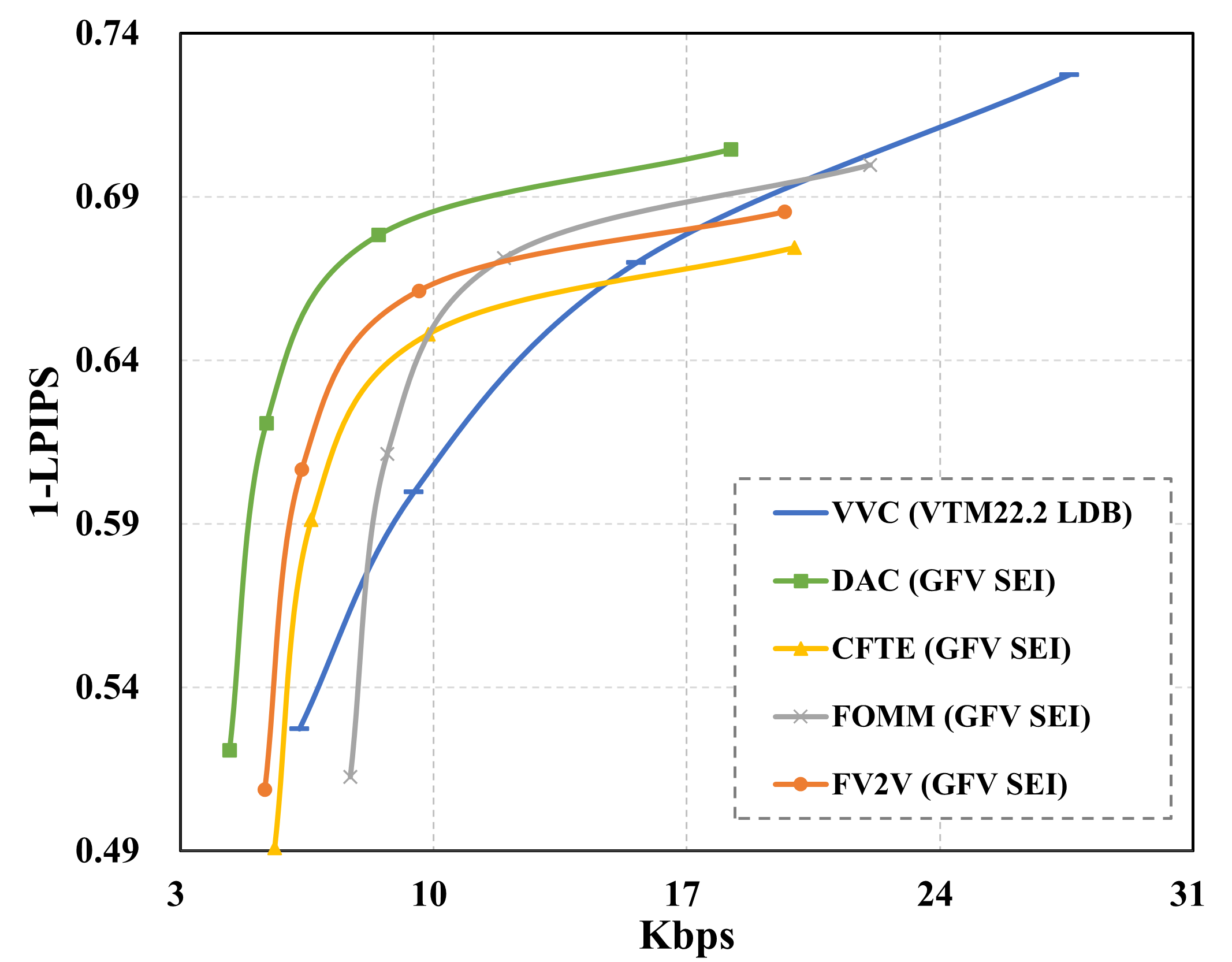}}
\caption{Rate-Distortion performance comparisons of VVC and four SEI-based GFVC approaches (CFTE, FV2V, DAC and FOMM) in terms of rate-DISTS and rate-LPIPS for both 256$\times$256 and 512$\times$512 resolutions. These experimental results are sourced from~\cite{chen2024standardizing}.} %
\label{standard_result}
\end{figure}

The SEI-based GFVC experimentation is conducted on JVET GFVC AhG software  implementations\footnote{\href{https://vcgit.hhi.fraunhofer.de/jvet/VVCSoftware_VTM/}{VTM Software with GFV SEI Implementation}}~\footnote{\href{https://vcgit.hhi.fraunhofer.de/jvet-ahg-gfvc/}{JVET AhG16 GFVC Software Tools}}, strictly following the JVET GFVC AhG test conditions~\cite{JVET-AJ2035}. As shown in Fig. \ref{standard_result}, standardizing GFVC with SEI messages can achieve promising RD performance when compared to the conventional VVC standard in higher bitrate ranges, particularly at a resolution of 256$\times$256. By elevating the face video resolution to 512$\times$512, even more significant coding performance advantages can be achieved across a wider bitrate ranges. These performance results demonstrate that SEI-based GFVC holds substantial potential in delivering efficient coding performance for face videos at or exceeding standard definition television resolutions.

\begin{figure}[t]
\centering
\subfloat[Architectural Optimization]{\includegraphics[width=0.25\textwidth]{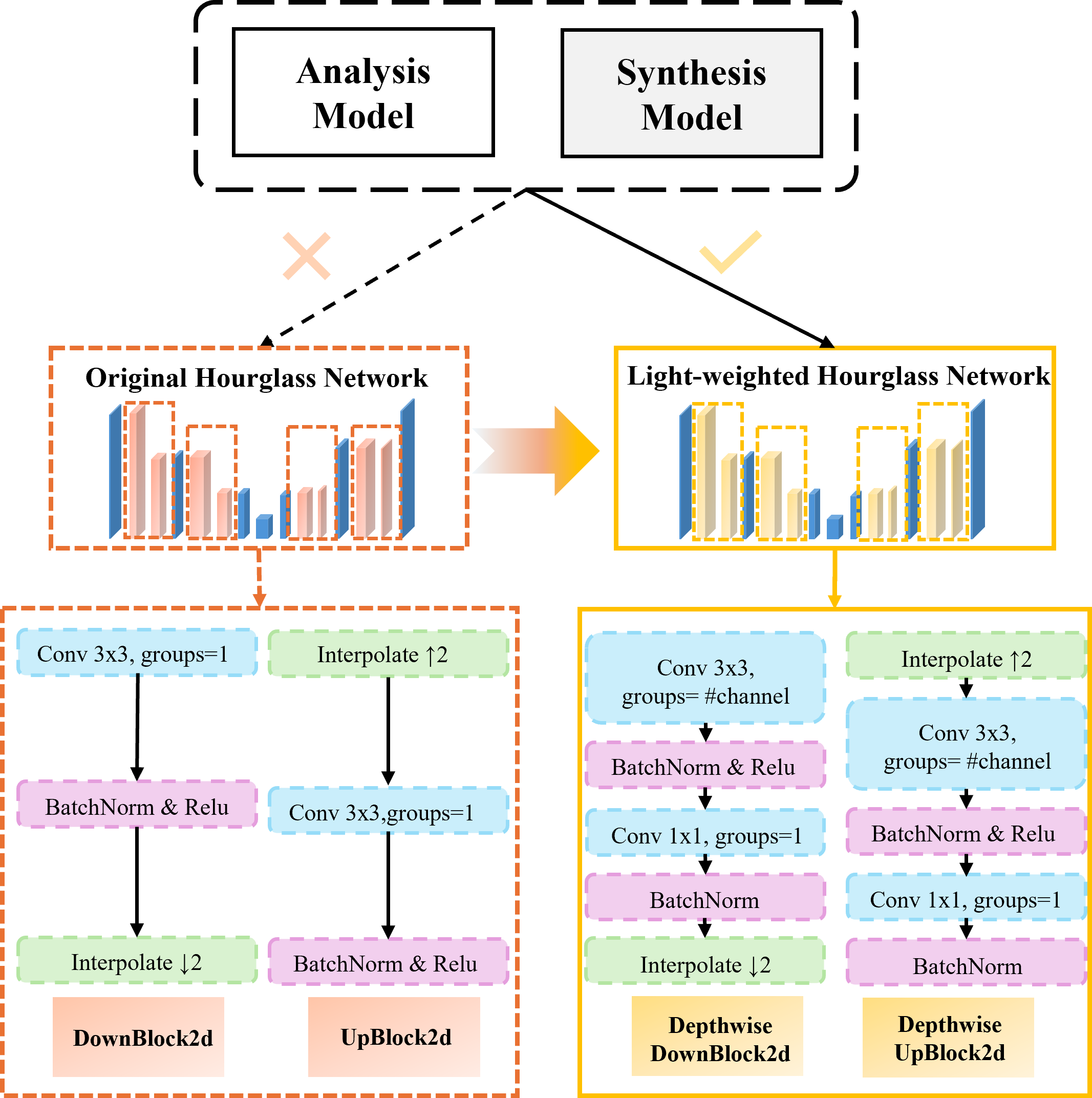}}
\hspace{0.02\textwidth}
\subfloat[Operational Optimization]
{\includegraphics[width=0.21\textwidth]{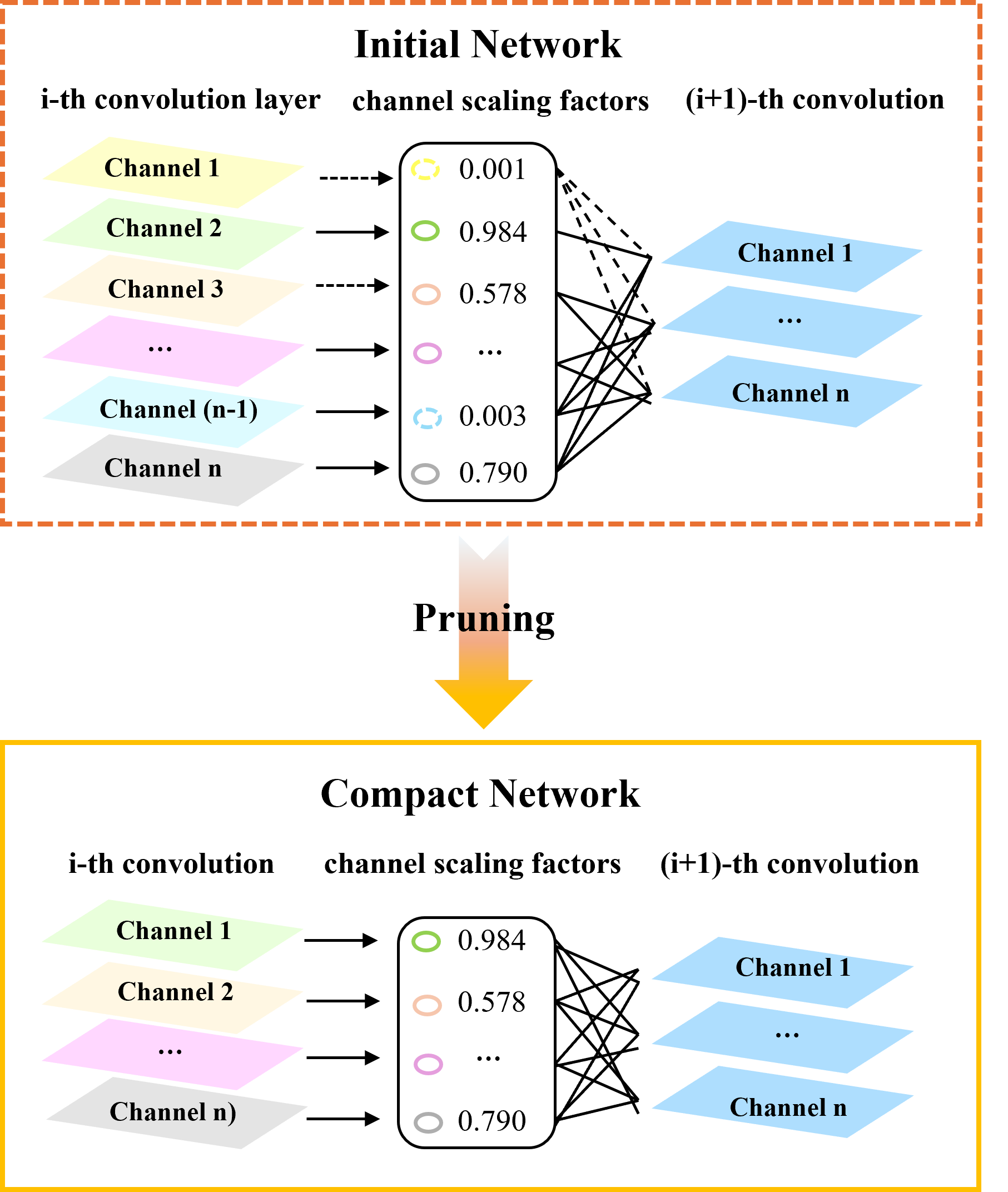}}
\caption{Illustrations of the proposed CFTE model with the architectural and operational optimizations.} 
\label{lightweight}
\end{figure}

\subsection{Model Lightweight for Low-complexity System Implementation}
\label{model_lightweight}
\subsubsection{Proposed Solution}
To enable future deployment on edge devices, we propose a lightweight approach integrating architectural and operational optimizations. 
Taking CFTE as an example, we illustrate our lightweight implementation in Fig.\ref{lightweight}

\textbf{Architectural Optimization:}  First, we replace some of the \( 3\times3\) convolutional layers in CFTE with depthwise separable convolutions~\cite{howard2017mobilenets}, which simplifies both the encoder and decoder. 
Depthwise separable convolution, which factorizes a standard convolution into a depthwise convolution and a \( 1 \times 1 \) pointwise convolution, is an efficient operation widely used in deep learning, particularly in convolutional neural networks. This approach can significantly reduce computational complexity and the number of network parameters without substantially compromising performance.

\textbf{Operational Optimization:}  Building on this architecture-level optimization framework, we further employ a channel pruning strategy guided by the scaling factors in Batch Normalization (BatchNorm) layers, inspired by the Network Slimming framework~\cite{liu2017learning}. Specifically, each BatchNorm layer following a convolutional layer contains a trainable scaling factor parameter $\gamma$, which scales the normalized activation values. The magnitude of the scaling factor $\gamma$ indicates the importance of the corresponding channel, and channels with smaller $\gamma$ values have less influence on subsequent layers. To optimize this, structured sparsity is introduced by imposing L1 regularization on $\gamma$ during training, where L1 regularization can promote sparsity in BatchNorm layers by driving the value of $\gamma$ toward zero.
The overall objective is defined as,
\begin{equation}
L_{\text{total}} = L_{\text{original}} + \lambda \sum_{\gamma \in \Gamma} |\gamma|,
\end{equation}
where $L_{\text{original}}$ represents the baseline loss of the CFTE model, $\Gamma$ denotes all BatchNorm scaling factors, and $\lambda$ controls the trade-off between task performance and channel sparsity.

After sparsity-regularized training, we obtain a $\gamma$-sparse network, where insignificant channels have smaller scaling factors. To better manage the model complexity and facilitate structural pruning, the network is first partitioned into distinct parts, each assigned a tailored pruning ratio \( r \), experimentally determined by its influence on model performance. 
For a given part with \( N \) BatchNorm layers, the number of channels to be pruned, denoted as \( k \), is calculated as \( k = N \times r \).
To perform pruning, we sort the \( \gamma \) values of the BatchNorm channels within each part in ascending order. 
The pruning threshold \( \gamma_{\text{threshold}} \) is defined as the \( k \)-th smallest value. Accordingly, the first \( k \) channels with the smallest \( \gamma \) values are removed, eliminating redundant channels while preserving those essential for the output.
After pruning, we fine-tune the compact network to recover potential accuracy losses caused by channel removal. This approach successfully identifies and removes redundant channels, significantly reducing computational complexity and model size while preserving performance.

Overall, the adopted lightweight techniques like architecture optimization and pruning operation can greatly reduce model parameters and inference complexity, which can be applied into all GFVC algorithms for their practical deployments. 

\subsubsection{Performance Analysis}
\begin{table}[t]
\renewcommand\arraystretch{1.35}
\centering
\caption{Comparison of model parameters and computational complexity between the original CFTE and the lightweight CFTE models}
\label{tab:comparison} 
\resizebox{0.5\textwidth}{!}{
\begin{tabular}{ccccccccc}
\toprule
& \multicolumn{3}{c}{Parameters (M)} & \multicolumn{3}{c}{Complexity (kMACs/pixel)} \\
\cmidrule(lr){2-4} \cmidrule(lr){5-7}
& Encoder & Decoder & Total & Encoder & Decoder & Total\\
\midrule
Original & 14.13 & 43.90 & 58.03 & 15.10 & 836.90 & 852.00\\
Lightweight & 1.36 & 4.24 & 5.60 & 0.82 & 124.96 & 125.78 \\
\midrule
Reduction Ratio & 90.37\% & 90.34\% & 90.35\% & 94.57\% & 85.07\% & 85.24\%\\
\bottomrule
\end{tabular}}
\end{table}

\begin{figure}[t]
\centering
\vspace{-1.3em}
\subfloat[Rate-DISTS]{\includegraphics[width=0.245\textwidth]{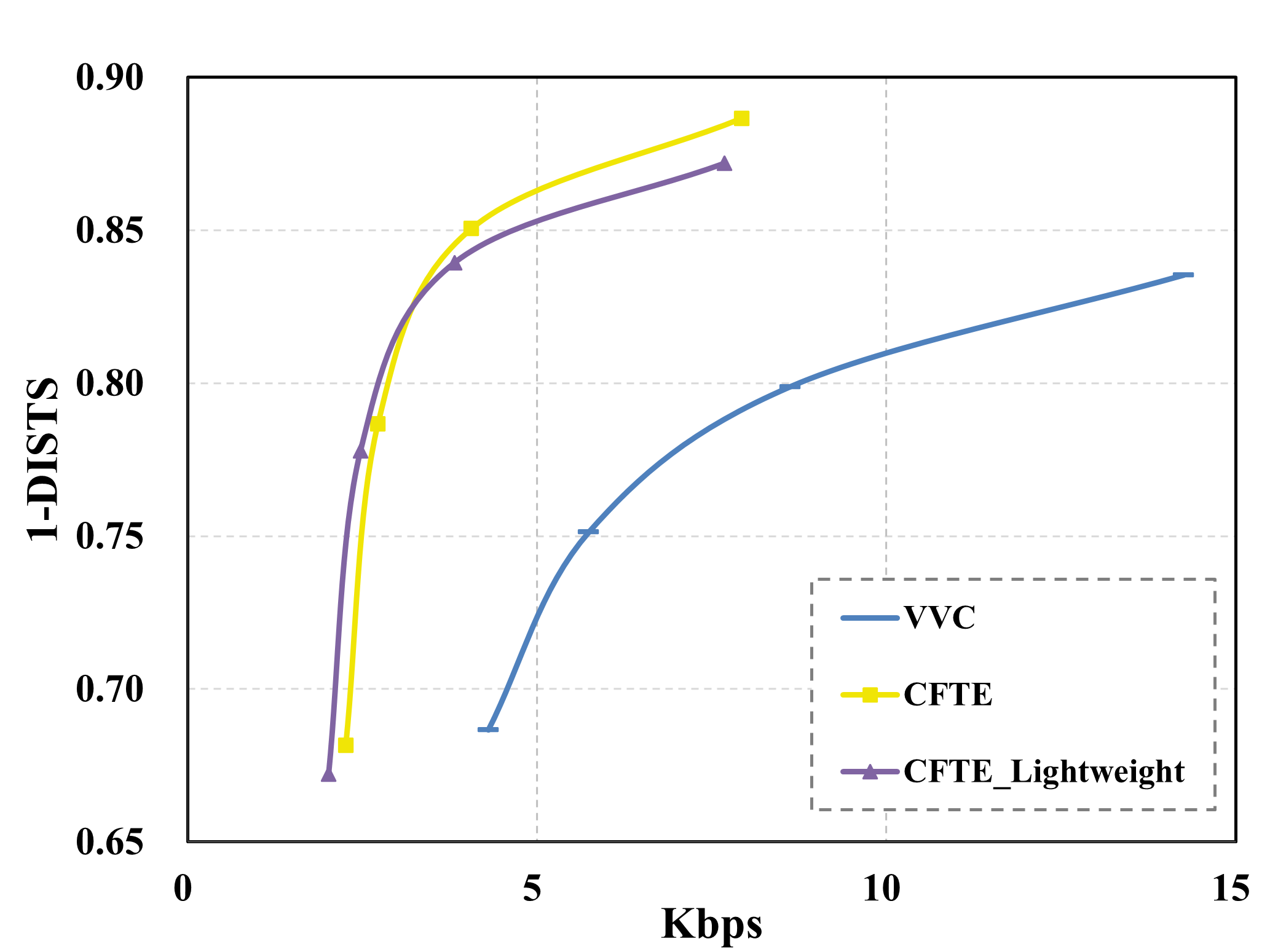}}
\subfloat[Rate-LPIPS]{\includegraphics[width=0.245\textwidth]{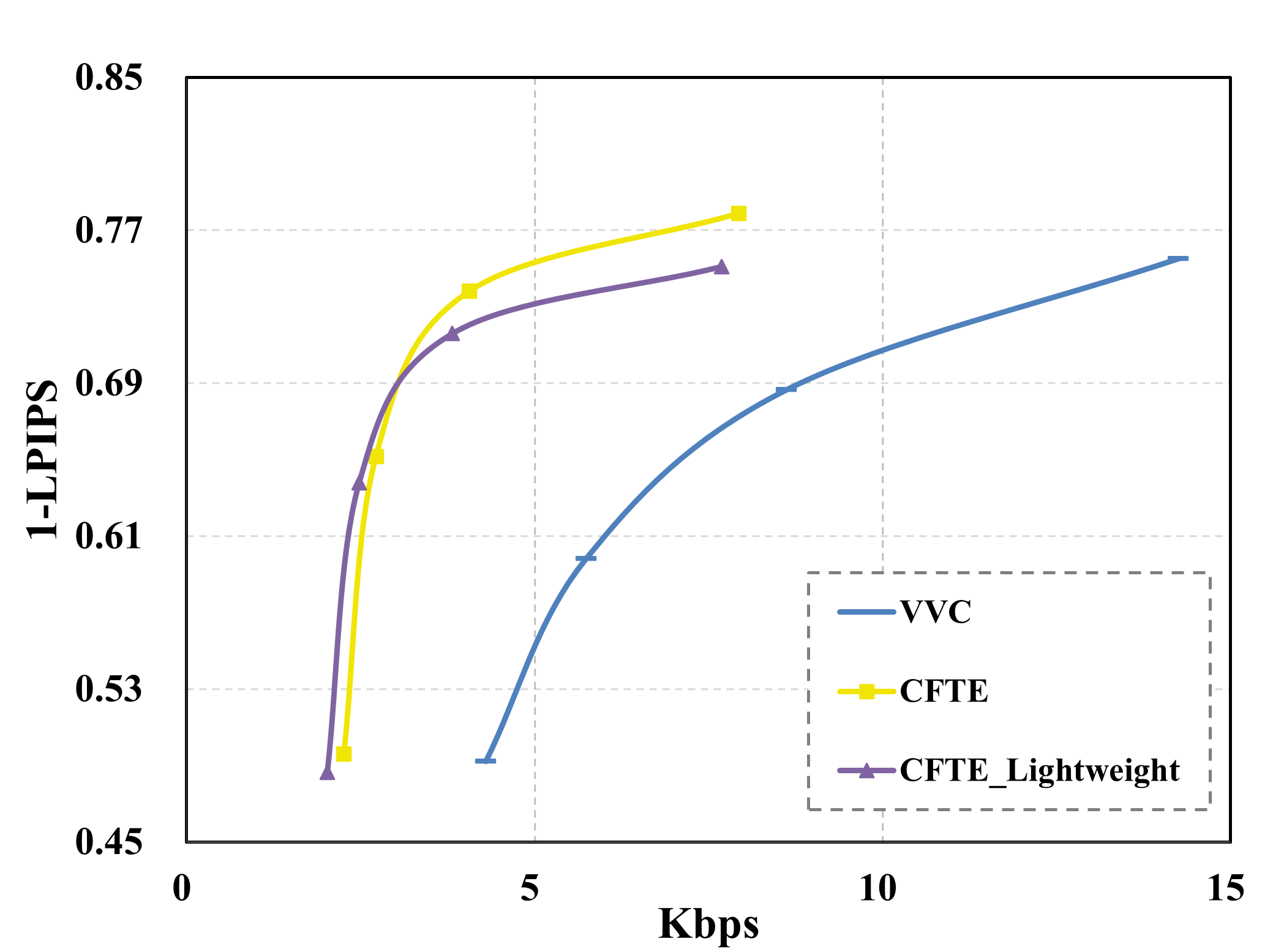}}
\caption{Rate-Distortion performance comparisons of VVC, CFTE and  CFTE\_Lightweight in terms of rate-DISTS and rate-LPIPS.} 
\label{RD}
\vspace{-0.5em}
\end{figure}


The training and testing of our lightweight CFTE model strictly follow experimental settings described on Section \ref{sec:CTC}. 
We evaluate the model lightweight efficiency in terms of number of parameters (M), computational complexity (kMACs/pixel) and RD performance.
As shown in Table \ref{tab:comparison}, the lightweight CFTE model significantly reduces total parameters by 90.35\% from 58.03 M to 5.60 M and computational complexity by 85.24\% from 852.00 to 125.78 kMACs per pixel.
Fig. \ref{RD} provides the RD performance comparisons among the VVC codec, the original CFTE model and the proposed lightweight CFTE model  in terms of DISTS and LPIPS.
After lightweight optimization, the proposed model does not result in noticeable performance drop compared with the original model, and both outperform the VVC codec at ultra-low bitrate ranges.
Overall, our proposed lightweight solution successfully reduces the number of parameters while maintaining competitive RD performance, highlighting its suitability for resource-constrained applications, such as mobile devices or embedded systems.

\begin{figure*}[tb]
\centering
\centerline{\includegraphics[width=1 \textwidth]{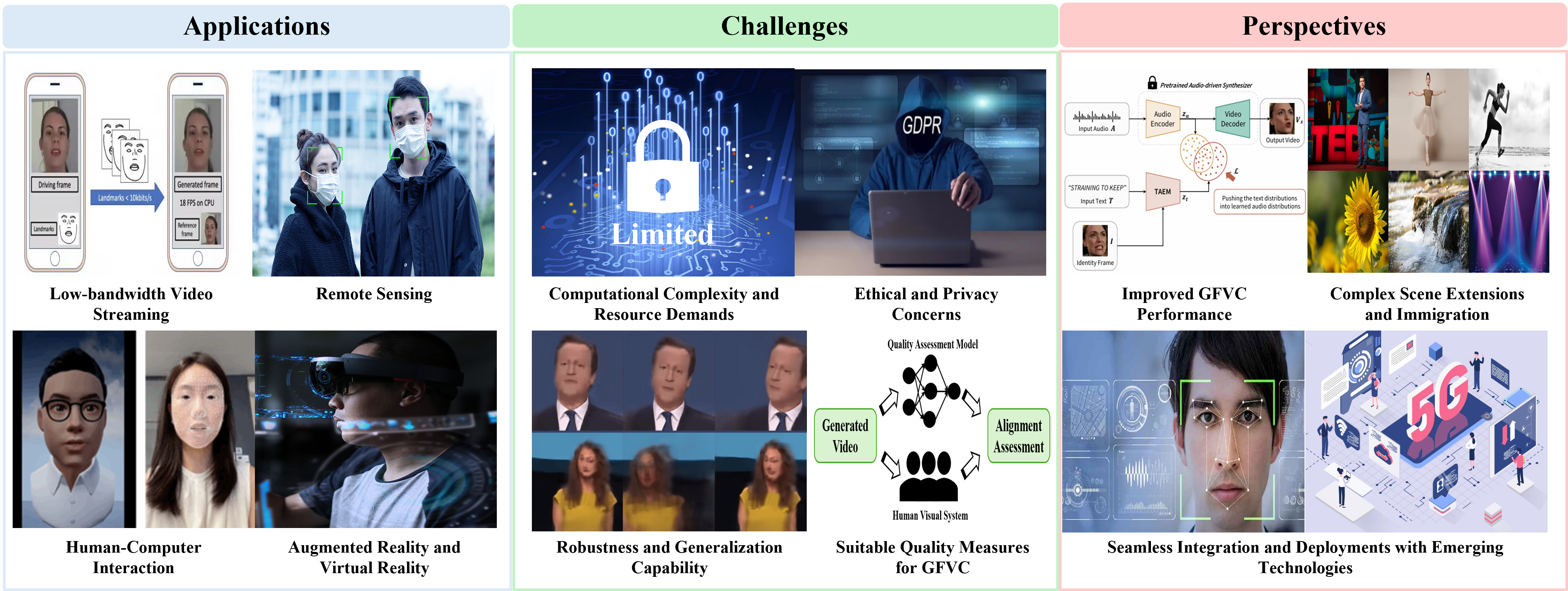}}  
\caption{Illustrations for potential applications, challenges, and  envisions in the domain of generative coding especially for GFVC techniques.}
\label{fig_section6} 
\end{figure*}

\section{Potential Applications, Challenges and Perspectives }
Potential applications, challenges, and envisions in the realm of generative coding especially for GFVC techniques are illustrated in Fig. \ref{fig_section6}. Applications span diverse fields like video streaming, remote sensing, and more. Challenges include computational demands, model interpretability, quality assessment and ethical concerns. Envisioned advancements involve improved GFVC compression performance, complex scene extension and seamless integration with emerging technologies.

\subsection{Applications}
GFVC aims to offer transformative potentials in diverse domains, including but not limited to:
\begin{itemize}
\item{\textbf{Low-bandwidth Video Streaming:} GFVC enables real-time face video communication in constrained bandwidth environments by leveraging deep generative models to reconstruct high-quality face videos. Furthermore, GFVC can cater to personalized content delivery and facial signal filtering according to individual user preferences or device capabilities.} 
\item{\textbf{Remote Sensing and Human-Computer Interaction:} GFVC is expected to support human-centered monitoring due to efficient transmission of facial data for identity verification or behavior analysis in surveillance systems. In addition, GFVC can facilitate expressive communication via digital assistants, chatbots, or virtual agents by reconstructing accurate facial expressions.} 
\item{\textbf{Augmented Reality (AR) and Virtual Reality (VR):} GFVC can play a pivotal role in offering immersive experiences within AR/VR applications by reconstructing realistic facial animations for avatars and facilitating telepresence in virtual environments. Moreover, GFVC contributes to the creation of lifelike digital characters in interactive media, enhancing gaming and media experiences through its capabilities.} 
\end{itemize}

\subsection{Challenges}
Despite its potential for applications ranging from telepresence to immersive media, GFVC faces several critical challenges that should be addressed to unlock its full potential. Below is a detailed discussion of these challenges.

\begin{itemize}
\item{\textbf{Computational Complexity and Resource Demands:}
GFVC relies on deep generative models (\textit{e.g.,} GANs, VAEs, diffusion models), which require significant computational resources for training and inference processes. Furthermore, the deployment of GFVC on devices with limited resources or at the network edge, like smartphones and Internet of Things (IoT) sensors, continues to pose significant challenges due to the high computational demands imposed by these models. Despite the model lightweight efforts were made as described in Section~\ref{model_lightweight}, it is necessary to optimize efficiency, enhance scalability, and facilitate the seamless integration of GFVC into edge computing environments for broader practical applications.} 

\item{\textbf{Model Interpretability and Reliability:} 
Understanding and explaining the behavior of deep generative models utilized in GFVC can well enhance trustworthiness of these GFVC models in critical applications. It is expected to lower the intricacies within these models, arising from complex training approaches and data representations.
Additionally, ensuring robustness against adversarial attacks and unintended biases is crucial for the safe deployment of GFVC systems. Proactive steps such as bias detection algorithms and rigorous model evaluations are crucial for promoting fairness and transparency in GFVC applications. } 

\item{\textbf{Quality Measures for Evaluation and Optimization:} GFVC usually operates within the feature domain, diverging from conventional pixel-level quality metrics like PSNR and SSIM commonly used in traditional video coding. While perceptual-based metrics such as DISTS and LPIPS have emerged to address feature-domain evaluation, they still fall short of adequately assessing the quality of generative content. In addition, human-centric applications (\textit{e.g.,} telemedicine and online meetings) need high fidelity in facial expressions and emotions, requiring rigorous subjective evaluations to avoid biases. Hence, there is a critical need to embrace tailor-made metrics to accurately measure performance efficacy and refine the algorithmic design for generative compression tasks.} 

\item{\textbf{Robustness and Generalization Capability:} Constrained by the inherent limitations of deep generative models, the current GFVC's reconstruction occasionally encounters visual artifacts and occluded distortions, leading to suboptimal visual quality experiences. This instability in reconstruction quality undermines the practical viability of applications to a certain degree. Consequently, there is a pressing need to delve into more sophisticated and resilient technologies built upon current algorithms to enhance the generalization capacity and robustness of GFVC models.} 

\item{\textbf{Ethical and Privacy Concerns:} 
GFVC models use diverse datasets for training, which may not always be readily available due to privacy concerns or data accessibility issues. Ensuring access to suitable training data while addressing privacy considerations presents a significant challenge in the development and deployment of GFVC systems.
In addition, GFVC’s ability to synthesize realistic face videos may raise ethical concerns, including misuse for misinformation or identity theft. Guaranteeing adherence to stringent data privacy laws such as the General Data Protection Regulation (GDPR) during the handling of facial data emerges as a pivotal concern, especially in contexts like surveillance and tele-medicine applications.} 
\end{itemize}

\subsection{Perspectives}
The field of GFVC is poised for significant growth, with several promising directions for future research and development:

\begin{itemize}
\item{\textbf{Improved GFVC Performance:}
Improving the compression performance of GFVC can involve a combination of algorithm improvement, architecture optimization and rate distortion optimization strategies. Here are some approaches that can be explored to boost the RD performance: 1) Designing and optimizing the architecture of advanced generative models (\textit{e.g.,} diffusion models and transformers) specifically for face video coding tasks. 2) Incorporating temporal information by using recurrent neural networks (RNNs), Long Short-Term Memory (LSTM) networks, or 3D convolutional neural networks to model the temporal dependencies in face video sequences. 3) Combining GFVC with other modalities (\textit{e.g.,} audio and text) for holistic compression and reconstruction of multimedia content. 4) Developing custom loss functions that are tailored to the specific requirements of face video coding, such as perceptual loss functions that consider the visual quality of the generated videos and align with human visual perception. and 5) Implementing optimized quantization and compression techniques (\textit{e.g.,} entropy coding, adaptive quantization, and transform coding) for more efficient latent space compression. } 

\item{\textbf{Complex Scene Extensions:}
When transitioning from GFVC to generative video coding for complex scenes beyond face video, it should entail integrating advanced techniques and enhance the model's scene understanding capabilities to encompass a wider range of objects, backgrounds, and interactions. Different from face video, human body videos~\cite{yin2024generative} or even natural dynamic videos~\cite{yin2025compressing} have higher requirements in terms of content complexity, scene interaction, and dynamic texture,  more technical challenges need to be overcome during the compression process. For example, by incorporating hierarchical representations (\textit{e.g.,} body pose estimation, clothing dynamics modeling and background information) or employing scene graphs to model relationships between scene elements (\textit{e.g.,} flowing water, moving trees, or changing lighting), future generative video coding techniques can capture intricate scene dynamics and diverse elements. As such, generative video coding is expected to be developed towards full-scenarios with high-quality, temporally consistent, and semantically rich video outputs. } 

\item{\textbf{Seamless Integration and Deployments with Emerging Technologies:}
To better ensure efficiency, security, and compatibility for GFVC integration and deployment, a series of cutting-edge technologies like real-time processing and cloud integration should be involved. Herein, the possible directions are listed: 1) 5G and Beyond: exploiting the low-latency and high bandwidth of 5G networks to enable real-time GFVC applications. 2) IoT Ecosystems: deploying lightweight GFVC models on IoT devices for efficient video surveillance, smart homes, and connected automotive systems. and 3) Energy-Efficient Models: developing lightweight generative architectures and hardware accelerators for efficient deployment on mobile or edge devices.} 

\end{itemize}

\section{Conclusion}
This paper presents a comprehensive survey on GFVC, highlighting its potentially transformative role in efficient face video communication through semantic-level representation and generative model-based coding.
We systematically review existing GFVC methods and provide a benchmark to evaluate their performance, demonstrating their superiority over the latest VVC codec.
In addition, we establish a large-scale GFVC database with human perception scores and uncover the appropriate indicators to evaluate GFVC-reconstructed video quality. 
Furthermore, we summarize the state-of-the-art GFVC standardization efforts, implement a low-complexity GFVC system, and discuss the possible applications, unsolved challenges and promising opportunities within this domain.
In general, this paper positions GFVC as a transformative technology for next-generation face video coding, where generative models transcend compression to enable bandwidth-intelligent communication. This comprehensive survey aims to serve as a foundational resource, propelling further explorations and developments in GFVC technologies.

\normalem
\bibliographystyle{IEEEtran}
\bibliography{main}

\vspace{-3em}
\begin{IEEEbiography}
[{\includegraphics[width=1in,height=1.25in,clip,keepaspectratio]{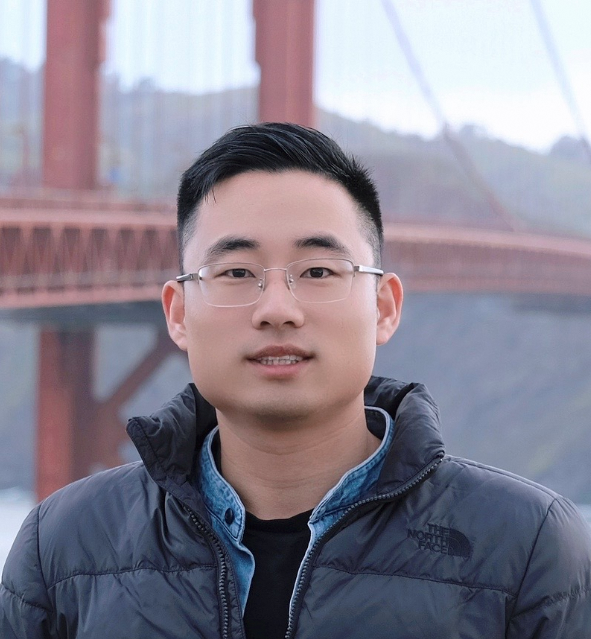}}]{Bolin Chen} received the B.S. degree in communication engineering from Fuzhou University in July 2020 and the Ph.D. degree in computer science from City University of Hong Kong in February 2025, respectively. He is currently a senior algorithm engineer at Alibaba DAMO Academy, and is also a joint postdoctoral fellow at Alibaba DAMO Academy and Fudan University. Prior to this, he was a research assistant and a short-term postdoctoral fellow at City University of Hong Kong. He has submitted more than 30 technical proposals to ISO/MPEG and ITU-T standards, and published more than 20 refereed journal articles/conference papers. His research interests include video compression, quality assessment and multimedia processing.
\end{IEEEbiography}

\vspace{-3em}
\begin{IEEEbiography}[{\includegraphics[width=1in,height=1.25in,clip,keepaspectratio]{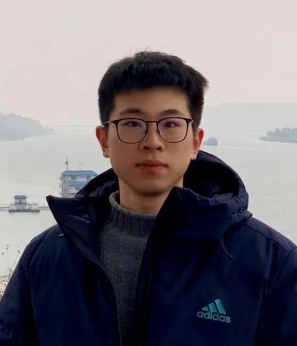}}]{Shanzhi Yin}  received the B.E. degree in communication engineering from Wuhan University of Technology, Wuhan, China, in 2020, and the M.S. degree in information and communication engineering from Harbin Institute of Technology, Shenzhen, China, in 2023. He is currently pursuing the Ph.D. degree with the Department of Computer Science, City University of Hong Kong. His research interests include video compression and generation.
\end{IEEEbiography}
\vspace{-3em}
\begin{IEEEbiography}
[{\includegraphics[width=1in,height=1.25in,clip,keepaspectratio]{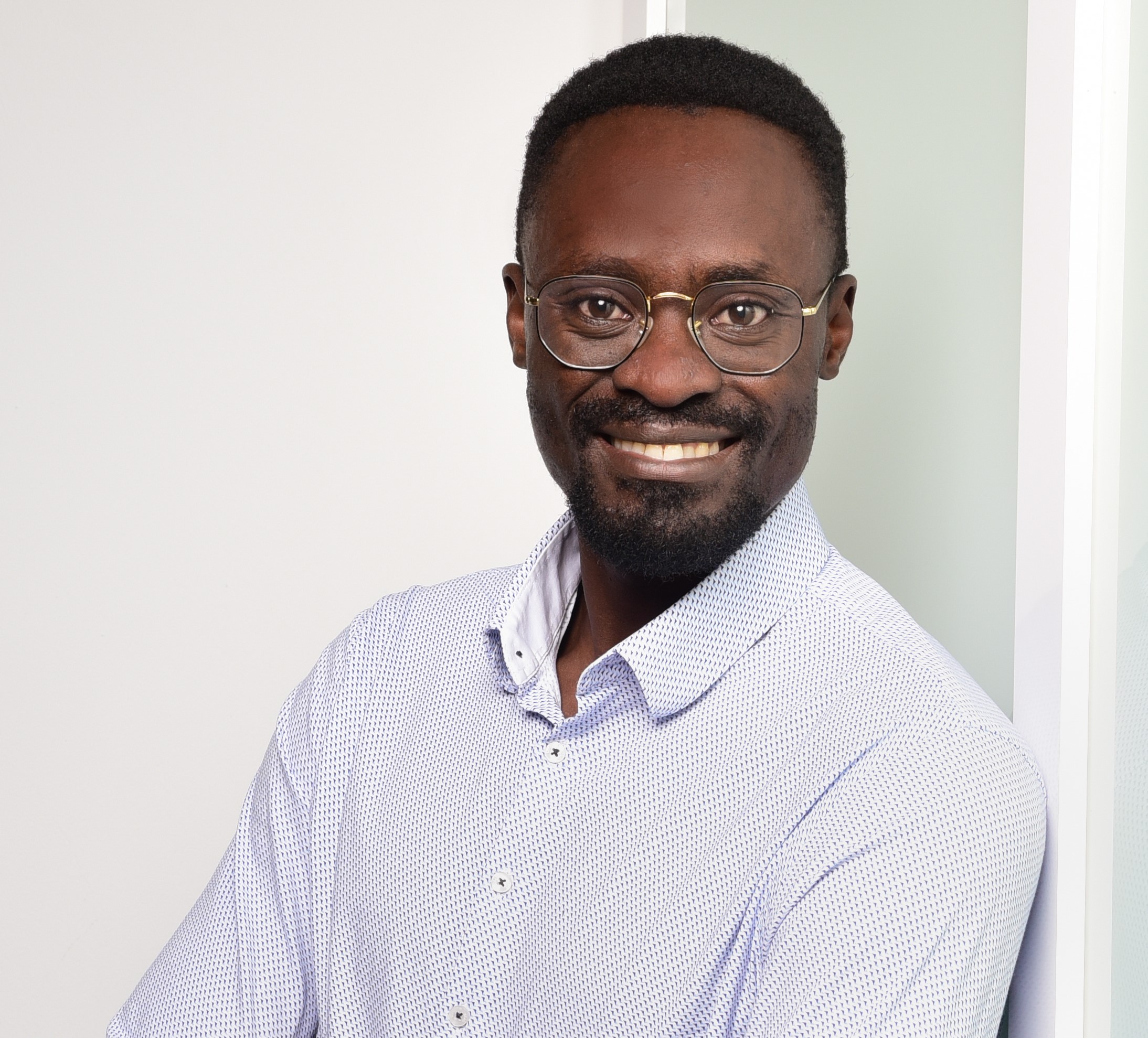}}]{Goluck Konuko}  received the B.S. degree in Electrical and Electronics Engineering from the Technical University of Kenya in December 2016, the M.Sc. degree in Multimedia Networking, and the Ph.D. degree in Signal Processing from Universit\'e Paris-Saclay in July 2020 and January 2025, respectively. He is currently a Research Scientist at Fraunhofer IIS - Erlangen, Germany. Prior to this, he worked as a Research Engineer at Telecom Paris, where he focused on optimization algorithms for real-time adaptive video streaming. His research interests encompass video coding for both human and machine vision, optimization techniques for real-time content delivery, subjective video quality prediction, and systems engineering.
\end{IEEEbiography}

\begin{IEEEbiography}
[{\includegraphics[width=1in,height=1.25in,clip,keepaspectratio]{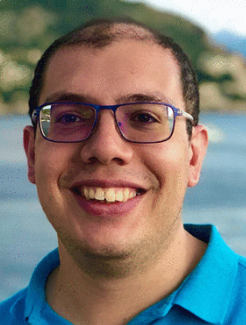}}]{Giuseppe Valenzise} is a CNRS researcher at Université Paris-Saclay, CNRS, Central-eSupélec, in the Laboratoire des Signaux et Systémes, where he is currently the head of the Multimedia and Networking team. He is the Editor in Chief of the EURASIP Journal on Image and Video Processing. Giuseppe obtained his Ph.D. degree from Politecnico di Milano, Italy, in 2011. He joined the French Centre National de la Recherche Scientifique (CNRS) in 2012. His research interests span different fields of image and video processing, including traditional and learning-based image and video compression, light fields and point cloud coding, image/ video quality assessment, high dynamic range imaging and applications of machine learning to image and video analysis. He has co-authored one book and over 130 research publications. He received the EURASIP Early Career Award in 2018 for significant contributions to video coding and analysis. Giuseppe serves/has served as Associate Editor for IEEE Transactions on Circuits and Systems for Video Technology, IEEE Transactions on Image Processing (outstanding editorial board member award in 2022 and 2023), Elsevier Signal Processing: Image communication. He is the Chair of the Multimedia Signal Processing (MMSP) technical committee of the IEEE Signal Processing Society. He is/was also an elected member of the Multimedia Systems and Applications (MSA) technical committee of the IEEE Circuits and Systems Society, of the MMSP and IVMSP technical committees of the IEEE Signal Processing Society, and of the Technical Area Committee on Visual Information Processing of EURASIP. Giuseppe is one of the general co-chairs of the IEEE Int. Conference on Multimedia\&Expo (ICME) 2025, held in Nantes, France.
\end{IEEEbiography}

\begin{IEEEbiography}
[{\includegraphics[width=1in,height=1.25in,clip,keepaspectratio]{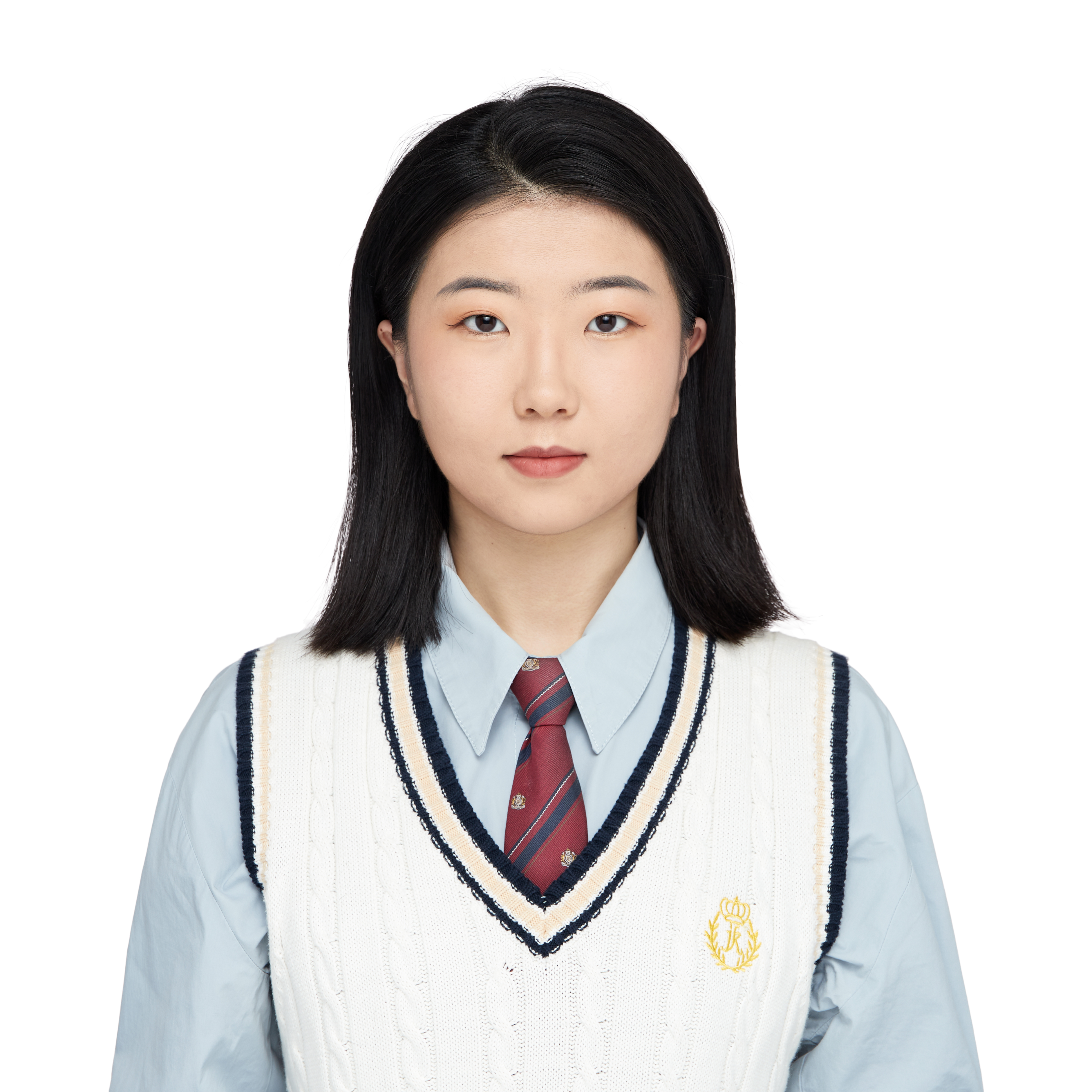}}]{Zihan Zhang} received the B.E. degree in information system from Harbin Institute of Technology, Weihai, China, in 2023. She is currently pursuing the Ph.D. degree with the Department of Computer Science, City University of Hong Kong, Hong Kong SAR, China. Her current research interests include video compression. 
\end{IEEEbiography}

\begin{IEEEbiography}[{\includegraphics[width=1in,height=1.25in,clip,keepaspectratio]{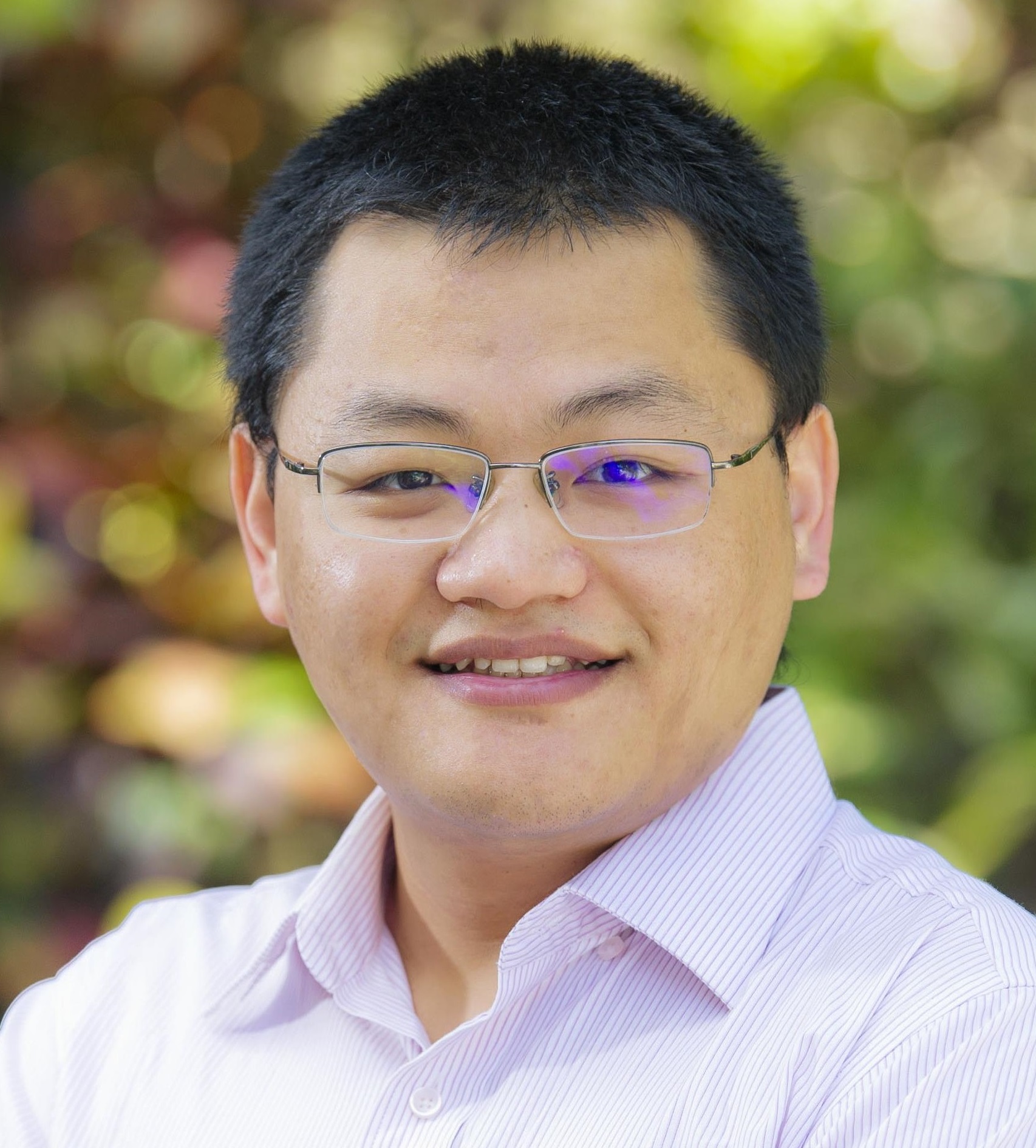}}]{Shiqi Wang} (Senior Member, IEEE) received the B.S. degree in computer science from the Harbin Institute of Technology in 2008 and the Ph.D. degree in computer application technology from Peking University in 2014. From 2014 to 2016, he was a Post-Doctoral Fellow with the Department of Electrical and Computer Engineering, University of Waterloo, Waterloo, ON, Canada. From 2016 to 2017, he was a Research Fellow with the Rapid-Rich Object Search Laboratory, Nanyang Technological University, Singapore. He is currently an Associate Professor with the Department of Computer Science, City University of Hong Kong. He has proposed more than 50 technical proposals to ISO/MPEG, ITU-T, and AVS standards, and authored or coauthored more than 300 refereed journal articles/conference papers. His research interests include video compression, image/video quality assessment, and image/video search and analysis. He received the Best Paper Award from IEEE VCIP 2019, ICME 2019, IEEE Multimedia 2018, and PCM 2017. His coauthored article received the Best Student Paper Award in the IEEE ICIP 2018. He was a recipient of the 2021 IEEE Multimedia Rising Star Award in ICME 2021. He serves as an Associate Editor for \textsc{IEEE Transactions on Circuits and Systems for Video Technology}, \textsc{IEEE Transactions on Image Processing}, \textsc{IEEE Transactions on Multimedia} and \textsc{IEEE Transactions on Cybernetics}. 
\end{IEEEbiography}

\begin{IEEEbiography}[{\includegraphics[width=1in,height=1.25in,clip,keepaspectratio]{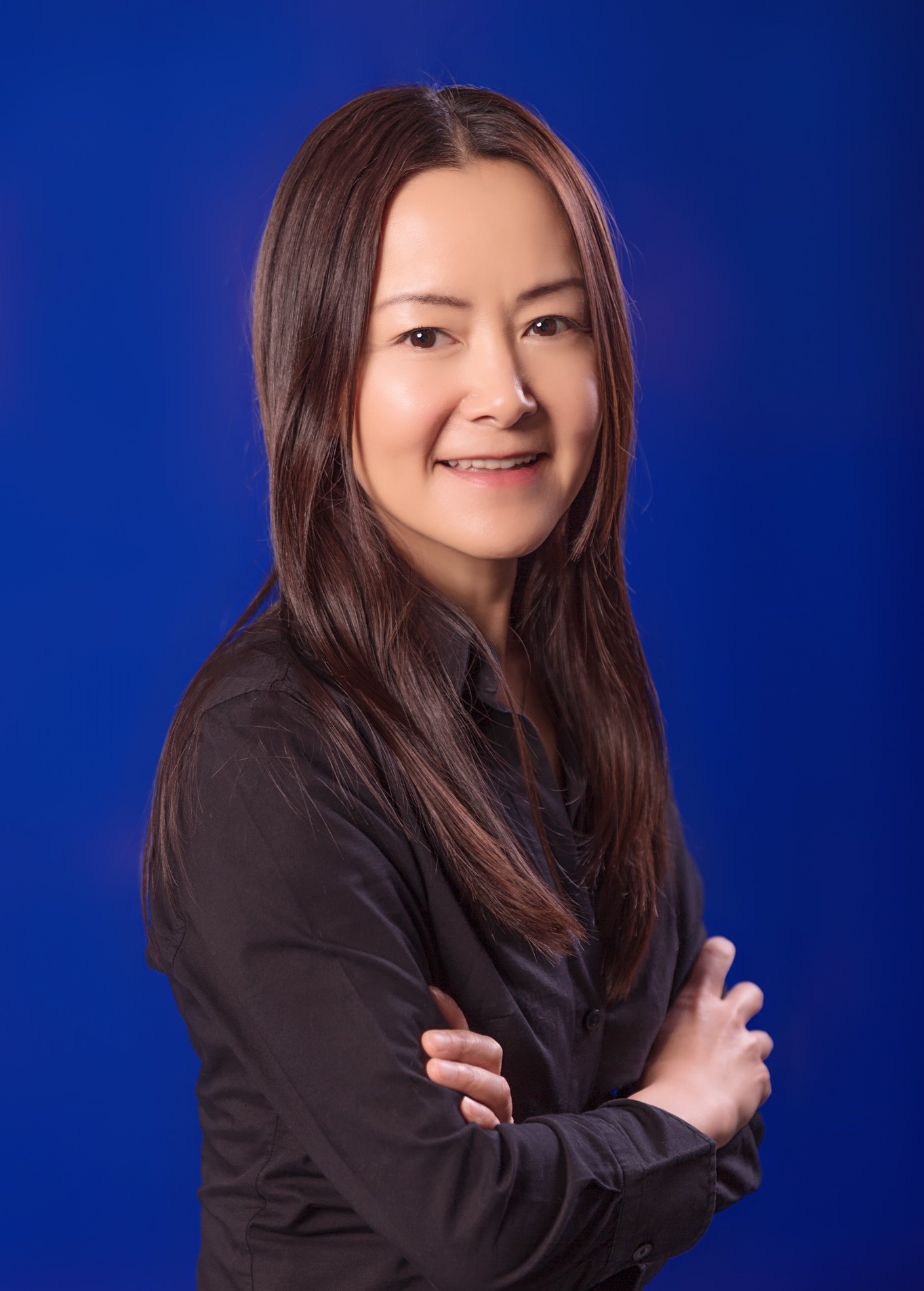}}]{Yan Ye} (Senior Member, IEEE) received the B.S. and M.S. degrees in electrical engineering from the University of Science and Technology of China in 1994 and 1997, respectively, and the Ph.D. degree in electrical engineering from the University of California at San Diego, in 2002. She is currently the head of Video Technology Lab at Alibaba DAMO Academy, Alibaba Group U.S., Sunnyvale, CA, USA, where she oversees multimedia standards development, video codec implementation, and AI-based video research. Prior to Alibaba, she was with the Research and Development Labs, InterDigital Communications, Image Technology Research, Dolby Laboratories, and Multimedia Research and Development and Standards, Qualcomm Technologies, Inc. She has been involved in the development of various video coding and streaming standards, including H.266/VVC, H.265/HEVC, scalable extension of H.264/MPEG-4 AVC, MPEG DASH, and MPEG OMAF. She has published more than 60 papers in peer-reviewed journals and conferences. Her research interests include advanced video coding, processing and streaming algorithms, real-time and immersive video communications, AR/VR, and deep learning-based video coding, processing, and quality assessment. 
\end{IEEEbiography}

\end{document}